\documentclass{article}

\usepackage[nonatbib, preprint]{neurips_2019}

\usepackage[utf8]{inputenc} 
\usepackage[T1]{fontenc}    
\usepackage{url}            
\usepackage{booktabs}       
\usepackage{amsfonts}       
\usepackage{nicefrac}       
\usepackage{microtype}      

\usepackage{color}
\usepackage{subfig}
\usepackage{bbm}
\usepackage{graphicx}
\usepackage{amsmath}
\usepackage{placeins} 
\usepackage{natbib}

\graphicspath{{figures/}}
\makeatletter 
\newcommand\footnoteref[1]{\protected@xdef\@thefnmark{\ref{#1}}\@footnotemark}
\makeatother

\newcommand{\iid}{\overset{i.i.d.}{\sim}}

\usepackage{times}

\title{Hyperparameter Learning via Distributional Transfer}

\author{
Ho Chung Leon Law \\
 University of Oxford\\
  \texttt{ho.law@stats.ox.ac.uk} \\ 
  \And 
  Peilin Zhao \\
  Tencent AI Labs\\
  \texttt{masonzhao@tencent.com} \\
  \And
  Lucian Chan \\
  University of Oxford\\
  \texttt{leung.chan@stats.ox.ac.uk} \\ 
  \And
  Junzhou Huang \\
  Tencent AI Labs\\
  \texttt{joehhuang@tencent.com} \\
  \And 
  Dino Sejdinovic \\
 University of Oxford\\
  \texttt{dino.sejdinovic@stats.ox.ac.uk} \\
}

\begin{document}

\maketitle

\begin{abstract}
Bayesian optimisation is a popular technique for hyperparameter learning but typically requires initial exploration even in cases where similar prior tasks have been solved. We propose to transfer information across tasks using learnt representations of training datasets used in those tasks. This results in a joint Gaussian process model on hyperparameters and data representations. Representations make use of the framework of distribution embeddings into reproducing kernel Hilbert spaces. The developed method has a faster convergence compared to existing baselines, in some cases requiring only a few evaluations of the target objective.
\end{abstract}

\section{Introduction}
Hyperparameter selection is an essential part of training a machine learning model and a judicious choice of values of hyperparameters such as learning rate, regularisation, or kernel parameters is what often makes the difference between an effective and a useless model. To tackle the challenge in a more principled way, the machine learning community has been increasingly focusing on Bayesian optimisation (BO) \citep{snoek2012practical}, a sequential strategy to select hyperparameters $\theta$ based on past evaluations of model performance. In particular, a Gaussian process (GP) \citep{rasmussen2004gaussian} prior is used to represent the underlying accuracy $f$ as a function of the hyperparameters $\theta$, whilst different acquisition functions $\alpha(\theta;f)$ are proposed to balance between exploration and exploitation. This has been shown to give superior performance compared to traditional methods \citep{snoek2012practical} such as grid search or random search. However, BO suffers from the so called `cold start' problem \citep{poloczek2016warm, swersky2013multi}, namely, initial observations of $f$ at different hyperparameters are required to fit a GP model. Various methods \citep{swersky2013multi, feurer2018scalable, springenberg2016bayesian, poloczek2016warm} were proposed to address this issue by transferring knowledge from previously solved tasks, however, 
initial random evaluations of the models are still needed to consider the similarity across tasks. This might be prohibitive: evaluations of $f$ can be computationally costly and our goal may be to select hyperparameters and deploy our model as soon as possible. We note that treating $f$ as a black-box function, as is often the case in BO, is ignoring the highly structured nature of hyperparameter learning -- it corresponds to training specific models on specific datasets. We make steps towards utilizing such structure in order to borrow strength across different tasks and datasets.

\textbf{Contribution.} We consider a scenario where a number of tasks have been previously solved and we propose a new BO algorithm, making use of the embeddings of the distribution of the training data \citep{blanchard2017domain, muandet2017kernel}. In particular, we propose a model that can jointly model all tasks at once, by considering an extended domain of inputs to model accuracy $f$, namely the distribution of the training data $\mathcal{P}_{XY}$, sample size of the training data $s$ and hyperparameters $\theta$. Through utilising \textit{all} seen evaluations from all tasks and meta-information, our methodology is able to learn a useful representation of the task that enables appropriate transfer of information to new tasks. As part of our contribution, we adapt our modelling approach to recent advances in scalable hyperparameter transfer learning \citep{perrone2018scalable} and demonstrate that our proposed methodology can scale linearly in the number of function evaluations. Empirically, across a range of regression and classification tasks, our methodology performs favourably at initialisation and has a faster convergence compared to existing baselines -- in some cases, the optimal accuracy is achieved in just a few evaluations.

\section{Related Work}
The idea of transferring information from different tasks in the context of hyperparameter learning has been studied in various settings \citep{swersky2013multi, feurer2018scalable, springenberg2016bayesian, poloczek2016warm, wistuba2018scalable, perrone2018scalable}. Amongst this literature, one common feature is that the similarity across tasks is captured only through the evaluations of $f$. This implies that sufficient evaluations from the task of interest is \textit{necessary}, before we can transfer information. This is problematic, if model training is computationally expensive and our goal is to employ our model as quickly as possible. Further, the hyperparameter search for a machine learning model in general is not a black-box function, as we have additional information available: the dataset used in training. In our work, we aim to learn feature representation of training datasets in-order to yield good initial hyperparameter candidates without having seen any evaluations from our target task.

While such use of such dataset features, called \emph{meta-features}, has been previously explored, current literature focuses on handcrafted meta-features\footnote{A comprehensive survey on meta-learning and handcrafted meta-features can be found in \citep[Ch.2]{automl_book}, \citep{feurer2015initializing}}. These strategies are not optimal, as these meta-features can be be very similar, while having very different $f$s, and vice versa. In fact a study on OpenML \citep{OpenML2013} meta-features have shown that the optimal set depends on the algorithm and data \citep{todorovski2000report}. This suggests that the reliance on these features can have an adverse effect on exploration, and we give an example of this in section \ref{sec:experiments}. To avoid such shortcomings, given the same input space, our algorithm is able to \emph{learn} meta-features directly from the data, avoiding such potential issues. Although \citep{kim2017learning} previously have also proposed to learn the meta-feature representations (for image data specifically), their proposed methodology requires the same set of hyperparameters to be evaluated for all previous tasks. This is clearly a limitation considering that different hyperparameter regions will be of interest for different tasks, and we would thus require excessive exploration of all those different regions under each task. To utilise meta-features, \citep{kim2017learning} propose to warm-start Bayesian optimisation \citep{gomes2012combining,reif2012meta,feurer2015initializing} by initialising with the best hyperparameters from previous tasks. This also might be sub-optimal as we neglect non-optimal hyperparameters that can still provide valuable information for our new task, as we demonstrate in section \ref{sec:experiments}. Our work can be thought of to be similar in spirit to \citep{klein2016fast}, which considers an additional input to be the sample size $s$, but do not consider different tasks corresponding to different training data distributions.

\section{Background}
\label{sec:background}
Our goal is to find:
\begin{equation*}
\theta^\ast_{\text{target}} = \text{argmax}_{\theta \in \Theta} f^{\text{target}}(\theta)
\end{equation*}
where $f^{\text{target}}$ is the target task objective we would like to optimise with respect to hyperparameters $\theta$. In our setting, we assume that there are $n$ (potentially) related source tasks $f^i, i=1,\dots n$, and for each $f^i$, we assume that we have $\{\theta^i_k, z^i_k\}_{k=1}^{N_i}$ from past runs, where $z^i_k$ denotes a noisy evaluation of $f^i(\theta^i_k)$ and $N_i$ denotes the number of evaluations of $f^i$ from task $i$. Here, we focus on the case that $f^i(\theta)$ is some standardised accuracy (e.g. test set AUC) of a trained machine learning model with hyperparameters $\theta$ and training data $D_i = \{\mathbf{x}^i_\ell, y^i_\ell\}_{\ell=1}^{s_i}$, where $\mathbf{x}^i_\ell \in \mathbb{R}^p$ are the covariates, $y^i_\ell$ are the labels and $s_i$ is the sample size of the training data. For a general framework, $D_i$ is any input to $f^i$ apart from $\theta$ (can be unsupervised) -- but following a typical supervised learning treatment, we assume it to be an i.i.d. sample from the joint distribution $\mathcal{P}_{XY}$.
For each task we now have: 
\begin{equation*}
(f^i, D_i = \{\mathbf{x}^i_\ell, y^i_\ell\}_{\ell=1}^{s_i}, \{\theta^i_k, z^i_k\}_{k=1}^{N_i}), \quad i=1,\dots n
\end{equation*}
Our strategy now is to measure the similarity between datasets (as a representation of the task itself), in order to transfer information from previous tasks to help us quickly locate $\theta^\ast_{\text{target}}$. In order to construct meaningful representations and measure between different tasks, we will make the assumption that $\mathbf{x}^i_\ell \in \mathcal{X}$ and $y^i_\ell \in \mathcal{Y}$ for all $i$, and that throughout the supervised learning model class is the same. While this setting might seem limiting, \citep{feurer2018scalable, poloczek2016warm} provides examples of many practical applications, including ride-sharing, customer analytics model, online inventory system and stock returns prediction. In all these cases, as new data becomes available, we might want to either re-train our model or re-fit our parameters of the system to adapt to a specific distributional data input. 


Intuitively, this assumption implies that the source of differences of $f^i(\theta)$ across $i$ and $f^{\text{target}}(\theta)$ is in the data $D_i$ and $D_{\text{target}}$. To model this, we will decompose the data $D_i$ into the joint distribution $\mathcal{P}^i_{XY}$ of the training data ($D_i =\{\mathbf{x}^i_\ell, y^i_\ell\}_{\ell=1}^{s_i} \overset{i.i.d.}\sim \mathcal{P}^i_{XY}$) and the sample size $s_i$ for task $i$. Sample size\footnote{Following \citep{klein2016fast}, in practice we re-scale $s$ to $[0,1]$, so that the task with the largest sample size has $s=1$.} is important here as it is closely related to model complexity choice which is in turn closely related to hyperparameter choice \citep{klein2016fast}. While we have chosen to model $D_i$ as $P^i_{XY}$ and $s_i$, in practice through simple modifications of the methodology we propose, it is possible to model $D_i$ as a set \citep{zaheer2017deep}. Under this setting, we will consider $f(\theta, \mathcal{P}_{XY}, s)$, where $f$ is a function on hyperparameters $\theta$, joint distribution $\mathcal{P}_{XY}$ and sample size $s$. For example, $f$ could be the negative empirical risk, i.e. 
\begin{equation*}
f(\theta, \mathcal{P}_{XY}, s) = -\frac{1}{s}\sum_{\ell=1}^s L(h_{\theta}(\mathbf{x}_\ell), y_\ell)),
\end{equation*} 
where $L$ is the loss function and $h_{\theta}$ is the model's predictor. To recover $f^i$ and $f^{\text{target}}$, we can evaluate at the corresponding $\mathcal{P}_{XY}$ and $s$, i.e. $f^{i}(\theta) = f(\theta, \mathcal{P}^{i}_{XY}, s_{i}), \; f^{\text{target}}(\theta) = f(\theta, \mathcal{P}^{\text{target}}_{XY}, s_{\text{target}}).$
In this form, we can see that similarly to assuming that $f$ varies smoothly as a function of $\theta$ in standard BO, this model also assumes smoothness of $f$ across $\mathcal{P}_{XY}$ as well as across $s$ following \citep{klein2016fast}. Here we can see that if two distributions and sample sizes are similar (with respect to a distance of their representations that we will learn), their corresponding values of $f$ will also be similar. In this source and target task setup, this would suggest we can selectively utilise information from previous source datasets evaluations $ \{\theta^i_k, z^i_k\}_{k=1}^{N_i}$ to help us model $f^{\text{target}}$. 

\section{Methodology}
\label{sec:method}
\subsection{Embedding of data distributions}
To model $\mathcal{P}_{XY}$, we will construct $\psi(D)$, a feature map on joint distributions for each task, estimated through its task's training data $D$. Here, we will follow similarly to \citep{blanchard2017domain} which considers transfer learning, and make use of kernel mean embedding to compute feature maps of distributions (cf. \citep{muandet2017kernel} for an overview). We begin by considering various feature maps of covariates and labels, denoting them by $\phi_x(\mathbf{x}) \in \mathbb{R}^a$, $\phi_y(y) \in \mathbb{R}^b$ and $\phi_{xy}([\mathbf{x}, y]) \in \mathbb{R}^c$, where $[\mathbf{x}, y]$ denotes the concatenation of covariates $\mathbf{x}$ and label $y$. 
Depending on the different scenarios, different quantities will be of interest.

\textbf{Marginal Distribution} $P_X$.   Modelling of the marginal distribution $P_X$ is useful, as we might expect various tasks to differ in the distribution of $\mathbf{x}$ and hence in the hyperparameters $\theta$, which, for example, may be related to the scales of covariates. We also might find that $\mathbf{x}$ is observed with different levels of noise across tasks. In this situation, it is natural to expect that those tasks with more noise would perform better under a simpler, more robust model (e.g. by increasing $\ell_2$ regularisation in the objective function). To embed $P_X$, we can estimate the kernel mean embedding $\mu_{P_X}$ \citep{muandet2017kernel} with $D$ by:
\vspace{-0.25cm}
\begin{equation*}
\psi(D) = \hat{\mu}_{P_X} = \frac{1}{s}\sum_{\ell=1}^{s} \phi_x(\mathbf{x}_\ell)
\end{equation*}
where $\psi(D) \in \mathbb{R}^a$ is an estimator of a representation of the marginal distribution $P_X$.

\textbf{Conditional Distribution $P_{Y|X}$}.  Similar to $P_X$, we can also embed the conditional distribution $P_{Y|X}$. This is an important quantity, as across tasks, the form of the signal can shift. For example, we might have a latent variable $W$ that controls the smoothness of a function, i.e. $P^i_{Y|X} = P_{Y|X, W=w_i}$. In a ridge regression setting, we will observe that those tasks (functions) that are less smooth would require a smaller bandwidth $\sigma$ in order to perform better. For regression, to model the conditional distribution, we will use the kernel conditional mean operator $C_{Y|X}$ \citep{song2013kernel} estimated with $D$ by:
\begin{eqnarray*} \hat{\mathcal{C}}_{Y|X} &=& \Phi_y^\top(\Phi_x\Phi_x^\top + \lambda I)^{-1}\Phi_x
 = \lambda^{-1} \Phi_y^\top( I - \Phi_x( \lambda I + \Phi_x^\top\Phi_x)^{-1}\Phi_x^\top)\Phi_x 
\end{eqnarray*}
where $\Phi_x = [\phi_x(\mathbf{x}_1), \dots, \phi_x(\mathbf{x}_{s})]^T \in \mathbb R^{s \times a}$, $\Phi_y =  [\phi_y(y_1), \dots, \phi_y(y_{s})]^T \in \mathbb R^{s \times b}$ and $\lambda$ is a regularisation parameter that we learn. It should be noted the second equality \citep{rasmussen2004gaussian} here allows us to avoid the $O(s^3)$ arising from the inverse. This is important, as the number of samples $s$ per task can be large. As $\hat{\mathcal{C}}_{Y|X} \in \mathbb{R}^{b\times a}$, we will flatten it to obtain $\psi(D) \in  \mathbb{R}^{ab}$ to obtain a representation of $P_{Y|X}$. In practice, as we rarely have prior insights into which quantity is useful for transferring hyperparameter information, we will model both the marginal and conditional distributions together by concatenating the two feature maps above. The advantage of such an approach is that the learning algorithm does not have to itself decouple the overall representation of training dataset into the information about marginal and conditional distributions which is likely to be informative.

\textbf{Joint Distribution $P_{XY}$}.  Taking an alternative and a more simplistic approach, it is also possible to model the joint distribution $P_{XY}$ directly. One approach is to compute the kernel mean embedding, based on concatenated samples $[\mathbf{x}, y]$, considering the feature map $\phi_{xy}$. Alternatively, we can also embed $\mathcal{P}_{XY}$ using the cross covariance operator $\mathcal{C}_{XY}$ \citep{gretton2015notes}, estimated by $D$ with:
\begin{equation*}
\hat{\mathcal{C}}_{XY} = \frac{1}{s}\sum_{\ell=1}^{s} \phi_x(\mathbf{x}_\ell) \otimes \phi_y(y_\ell) = \frac{1}{s}\Phi_x^\top \Phi_y \in \mathbb{R}^{a\times b}.
\end{equation*}
where $\otimes$ denotes the outer product and similarly to $\mathcal{C}_{Y|X}$, we will flatten it to obtain $\psi(D)\in \mathbb{R}^{ab}$. 


An important choice when modelling these quantities is the form of feature maps $\phi_x$, $\phi_y$ and $\phi_{xy}$, as these define the corresponding features of the data distribution we would like to capture. For example $\phi_x(\mathbf{x}) = \mathbf{x}$ and $\phi_x(\mathbf{x}) = \mathbf{x}\mathbf{x}^\top$ would be capturing the respective mean and second moment of the marginal distribution $P_x$. However, instead of defining a fixed feature map, here we will opt for a flexible representation, specifically in the form of neural networks (NN) for $\phi_x$, $\phi_y$ and $\phi_{xy}$ (except $\phi_y$ for classification\footnote{For classification, we use $\hat{\mathcal{C}}_{XY}$ and a one-hot encoding for $\phi_y$ implying a marginal embedding per class.}), in a similar fashion to \citep{wilson2016deep}. To provide a better intuition on this choice, suppose we have two task $i,j$ and that $\mathcal{P}_{XY}^i \approx \mathcal{P}_{XY}^j$ (with the same sample size $s$). This will imply that $f^i \approx f^j$, and hence $\theta^\ast_i \approx \theta^\ast_j$. However, the converse does not hold in general: $f^i \approx f^j$ does \textit{not} necessary imply $\mathcal{P}_{XY}^i \approx \mathcal{P}_{XY}^j$. For example, regularisation hyperparameters of a standard machine learning model are likely to be robust to rotations and orthogonal transformations of the covariates (leading to a different $P_X$). Hence, it is important to define a versatile model for $\psi(D)$, which can yield representations invariant to variations in the training data irrelevant for hyperparameter choice.

\subsection{Modelling $f$}
Given $\psi(D)$, we will now construct a model for $f(\theta, \mathcal{P}_{XY}, s)$, given observations $\left\{\{(\theta^i_k, \mathcal{P}^i_{XY}, s_i), z^i_k\}_{k=1}^{N_i}\right\}_{i=1}^n$, along with any observations on the target. Note that we will interchangeably use the notation $f$ to denote the model and the underlying function of interest. We will now focus on the algorithms distGP and distBLR, with additional details to be found in Appendix \ref{app:algorithm}.

\textbf{Gaussian Processes (distGP)}. We proceed similarly to standard BO \citep{snoek2012practical} using a GP to model $f$ and a normal likelihood (with variance $\sigma^2$ across all tasks\footnote{For different noise levels across tasks, we can allow for different $\sigma_i^2$ per task $i$ in distGP and distBLR.}) for our observations $z$,
\begin{equation*}
f \sim  GP(\mu, C)
\quad \quad z|\gamma  \sim \mathcal{N}(f(\gamma),\sigma^2)
\end{equation*}
where here $\mu$ is a constant, $C$ is the corresponding covariance function on $(\theta, \mathcal{P}_{XY}, s)$ and $\gamma$ is a particular instance of an input.
In order to fit a GP with inputs $(\theta, \mathcal{P}_{XY}, s)$, we use the following $C$:
\begin{equation*}
C(\{\theta_1, \mathcal{P}^1_{XY}, s_1\}, \{\theta_2, \mathcal{P}^2_{XY}, s_2\}) = \nu k_\theta(\theta_1, \theta_2) k_p([\psi(D_1), s_1],[\psi(D_2), s_2])
\end{equation*}
where $\nu$ is a constant, $k_\theta$ and $k_p$ is the standard Mat\'ern-$3/2$ kernel (with separate bandwidths across the dimensions). For classification, we additionally concatenate the class size ratio per class, as this is not captured in $\psi(D_i)$. Utilising $\left\{\{(\theta^i_k, \mathcal{P}^i_{XY}, s_i), z^i_k\}_{k=1}^{N_i}\right\}_{i=1}^n$, we can optimise  $\mu$, $\nu$, $\sigma^2$ and any parameters in $\psi(D)$, $k_\theta$ and $k_p$ using the marginal likelihood of the GP (in an end-to-end fashion).

\textbf{Bayesian Linear Regression (distBLR)}. While GP with its well-calibrated uncertainties have shown superior performance in BO \citep{snoek2012practical}, it is well known that they suffer from $O(N^3)$ computational complexity \citep{rasmussen2004gaussian}, where $N$ is the total number of observations. In this case, as $N=\sum_{i=1}^n N_i$, we might find that the total number of evaluations across all tasks is too large for the GP inference to be tractable or that the computational burden of GPs outweighs the cost of computing $f$ in the first place. To overcome this problem, we will follow \citep{perrone2018scalable} and use Bayesian linear regression (BLR), which scales linearly in the number of observations, with the model given by
\begin{equation*}
z|\beta \sim \mathcal{N}(\Upsilon \beta,\sigma^2 I )
\quad \quad  \beta \sim  \mathcal{N}(0, \alpha I) 
\quad \quad \Psi_i = [\psi(D_i), s_i]
\end{equation*}
\begin{equation*}
\Upsilon = [\upsilon([\theta^1_1, \Psi_1]), \dots, \upsilon([\theta^1_{N_1}, \Psi_1]), \dots, \upsilon([\theta^n_{1}, \Psi_n
]), \dots, \upsilon([\theta^n_{N_n}, \Psi_n
])]^\top \in \mathbb{R}^{N\times d}
\end{equation*}
where $\alpha > 0$ denotes the prior regularisation, and $[\cdot,\cdot]$ denotes concatentation. Here $\upsilon$ denotes a feature map on concatenated hyperparameters $\theta$, data embedding $\psi(D)$ and sample size $s$. Following \citep{perrone2018scalable}, we also employ a neural network for $\upsilon$.  
While conceptually similar to \citep{perrone2018scalable} who fits a BLR per task, here we consider a single BLR fitted jointly on all tasks, highlighting differences across tasks using meta-information available. The advantage of our approach is that for a given new task, we are able to utilise directly all previous information and one-shot predict hyperparameters without seeing \textit{any} evaluations from the target task. This is especially important when our goal might be to employ our system with only a few evaluations from our target task. In addition, a separate target task BLR is likely to be poorly fitted given only a few evaluations. Similar to the GP case, we can optimise $\alpha, \beta, \sigma^2$ and any unknown parameters in $\psi(D), \upsilon([\theta, \Psi])$ using the marginal likelihood of the BLR.
\subsection{Hyperparameter learning}
\label{sec:hyperlearn}
Having constructed a model for $f$ and optimised any unknown parameters through the marginal likelihood, in order to construct a model for the  $f^{\text{target}}$, we let $f^{\text{target}}(\theta)=f(\theta, \mathcal{P}^{\text{target}}_{XY}, s_{\text{target}})$. Now, to propose the next $\theta^{\text{target}}$ to evaluate, we can simply proceed with Bayesian optimisation on $f^{\text{target}}$, i.e. maximise the corresponding acquisition function $\alpha(\theta;f^{\text{target}})$. While we adopt standard BO techniques and acquisition functions here, note that the generality of the developed framework allows it to be readily combined with many advances in the BO literature, e.g. \cite{lobato2014, oh2018bock, mcleod_optimization_2018, snoek2012practical, wang2016parallel}.

\textbf{Acquisition Functions}. For the form of the acquisition function $\alpha(\theta;f^{\text{target}})$, we will use the popular expected improvement (EI) \citep{movckus1975bayesian}. However, for the first iteration, EI 
is not appropriate in our context, as these acquisition functions can favour $\theta$s with high uncertainty.
Recalling that our goal is to quickly select `good' hyperparameters $\theta$ with few evaluations, for the first iteration we will maximise the lower confidence bound (LCB)\footnote{Note this is not the upper confidence bound, as we want to \textit{exploit} and obtain a good starting initialisation.}, as we want to penalise uncertainties and exploit our knowledge from source task's evaluations. While this approach works well for the GP case, for BLR, we will use the LCB restricted to the best hyperparameters from previous tasks, as BLR with a NN feature map does not extrapolate as well as GPs in the first iteration. For the exact forms of these acquisition functions, implementation and alternative warm-starting approaches, please refer to Appendix \ref{app:warm-start}.

\textbf{Optimisation}.  We make use of ADAM \citep{kingma2014adam} to maximise the marginal likelihood until convergence. To ensure relative comparisons, we standardised each task's dataset features to have mean $0$ and variance $1$ (except for the unsupervised toy example), with regression labels normalised individually to be in $[0,1]$. As the sample size per task $s_i$ is likely to be large, instead of using the full set of samples $s_i$ to compute $\psi(D_i)$, we will use a different random sub-sample 
of batch-size $b$ for each iteration of optimisation. In practice, this parameter $b$ is dependent on the number of tasks, and the evaluation cost of $f$. It should be noted that a smaller batch-size $b$ would still provide an unbiased estimate of $\psi(D_i)$ 
At testing time, it is also possible to use a sub-sample of the dataset to avoid any computational costs arising from a large $\sum_i s_i$. When retraining, we will initialise from the previous set of parameters, hence few gradient steps are required before convergence occurs.

\textbf{Extension to other data structures.} Throughout the paper, we focus on examples with $\mathbf{x} \in \mathbb{R}^{p}$. However our formulation is more general, as we only require the corresponding feature maps to be defined on individual covariates and labels. For example, image data can be modelled by taking $\phi_x(\mathbf{x})$ to be a representation given by a convolutional neural network (CNN)\footnote{This is similar to \citep{law2018bayesian} who embeds distribution of images using a pre-trained CNN for distribution regression.}, while for text data, we might construct features using Word2vec \citep{mikolov2013distributed}, and then retrain these representations for hyperparameter learning setting. More broadly, we can initialize $\psi(D)$ to any meaningful representation of the data, believed to be useful to the selection of $\theta^\ast_{\text{target}}$. Of course, we can also choose $\psi(D)$ simply as a selection of handcrafted meta-features \citep[Ch. 2]{automl_book}, in which case our methodology would use these representations to measure similarity between tasks, while performing feature selection \citep{todorovski2000report}. In practice, learned feature maps via kernel mean embeddings can be used in conjunction with handcrafted meta-features, letting data speak for itself. In Appendix \ref{app:manual}, we provide a selection of $13$ handcrafted meta-features that we employ as baselines for the experiments below. 
\section{Experiments}
\label{sec:experiments}
\begin{figure}[t]
    \centering
    \includegraphics[scale=0.32]{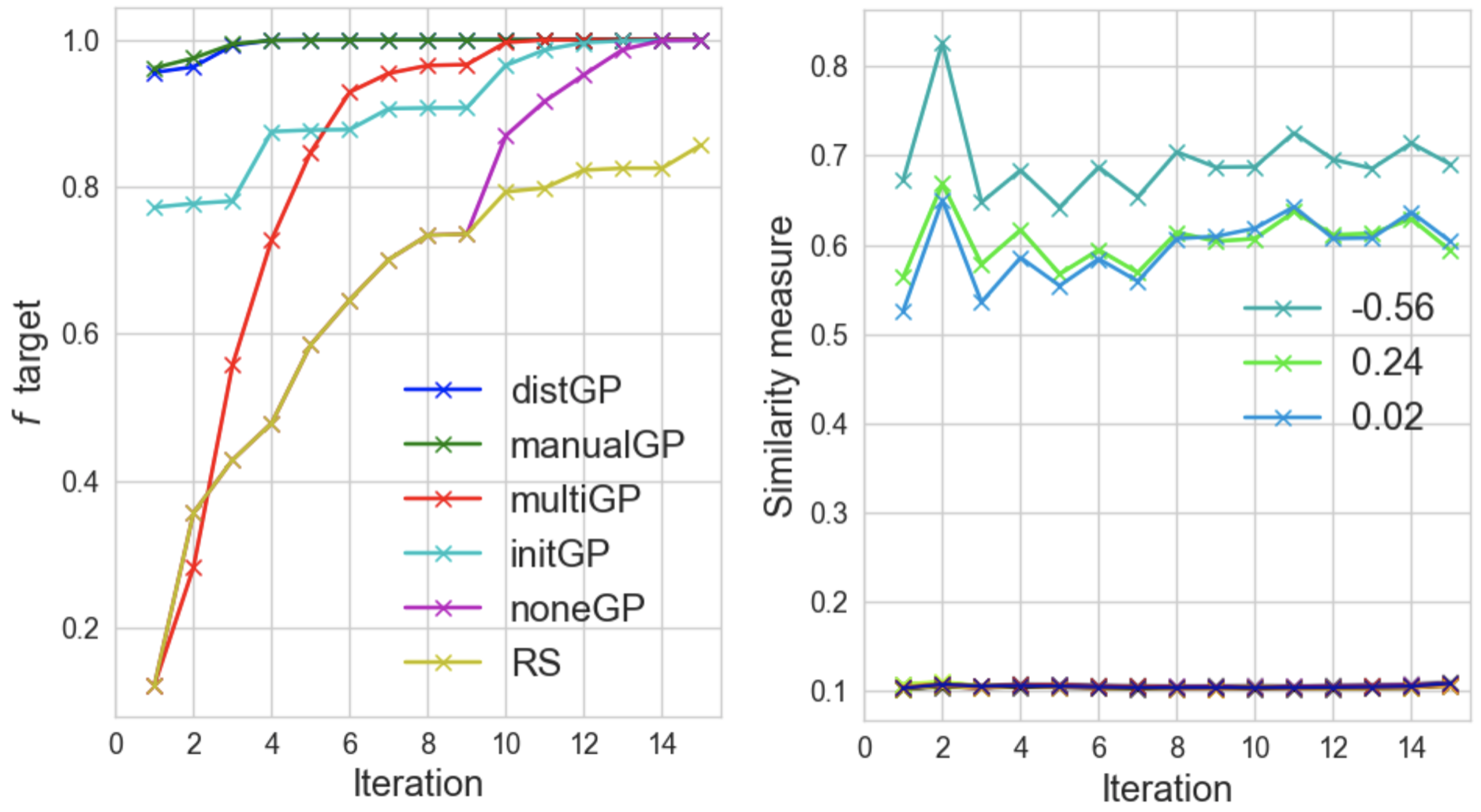}
    \caption{Unsupervised toy task over $30$ runs. \textbf{Left}: Mean of the \textit{maximum observed} $f^{target}$ so far (including any initialisation). \textbf{Right:} Mean of the similarity measure $k_p(\psi(D_i), \psi(D_{\text{target}}))$ for distGP. For clarity purposes, the legend \textit{only} shows the $\mu^i$ for the $3$ source tasks that are similar to the target task with $\mu^i=-0.25$. It is noted the rest of the source task have $\mu^i\approx 4$.}
    \label{fig:toy_sim}
\end{figure}
We will denote our methodology distBO, with BO being a placeholder for GP and BLR versions. For $\phi_x$ and  $\phi_y$ we will use a single hidden layer NN with $\tanh$ activation (with 20 hidden and 10 output units), except for classification tasks, where we use a one-hot encoding for $\phi_y$. For clarity purposes, we will focus on the approach where we separately embed the marginal and conditional distributions, before concatenation. Additional results for embedding the joint distribution can be found in Appendix  \ref{app:concat_embed}. For BLR, we will follow \citep{perrone2018scalable} and take feature map $\upsilon$ to be a NN with three 50-unit layers and $\tanh$ activation. For baselines, we will consider: 1) manualBO with $\psi(D)$ as the selection of $13$ handcrafted meta-features; 2) multiBO, i.e. multiGP \citep{swersky2013multi} and multiBLR \citep{perrone2018scalable} where no meta-information is used, i.e. task is simply encoded by its index (they are initialised with $1$ random iteration); 3) initBO \citep{feurer2015initializing} with plain Bayesian optimisation, but warm-started with the top $3$ hyperparameters, from the three most similar source tasks, computing the similarity with the $\ell_2$ distance on handcrafted meta-features;  4) noneBO denoting the plain Bayesian optimisation \citep{snoek2012practical}, with no previous task information; 5) RS denoting the random search. In all cases, both GP and BLR versions are considered. 

We use $\textit{TensorFlow}$ \citep{abadi2016tensorflow} for implementation, repeating each experiment $30$ times, either through re-sampling or re-splitting the train/test partition. For testing, we use the same number of samples $s_i$ for toy data, while using a 60-40 train-test split for real data. We take the embedding batch-size\footnote{Training time is less than $2$ minutes on a standard 2.60GHz single-core CPU in all experiments.} $b=1000$, and learning rate for ADAM to be 0.005. To obtain $\{\theta^i_k, z^i_k\}_{k=1}^{N_i}$ for source task $i$, we use noneGP to simulate a realistic scenario. Additional details on these baselines and implementation can be found in Appendix B and \ref{app:experiment}, with additional toy (\textit{non-similar source tasks scenario}) and real life (\textit{Parkinson's dataset}) experiments to be found in Appendix \ref{app:unseen_task} and \ref{app:parkinson}.

\subsection{Toy example.} To understand the various characteristics of the different methodologies, we first consider an "unsupervised" toy 1-dimensional example, where the dataset $D_i$ follows the generative process for some fixed $\gamma^i$:
$\mu^i \sim \mathcal{N}(\gamma^i,1); \ \{x^i_\ell\}_{\ell=1}^{s_i}|\mu^i \iid \mathcal{N}(\mu^i,1)$.
We can think of $\mu^i$ as the (unobserved) relevant property varying across tasks, and the unlabelled dataset as $D_i=\{x^i_\ell \}_{\ell=1}^{s_i}$. Here, we will consider the objective $f$ given by:
\begin{equation*}
f(\theta; D_i) = \exp\left(-\frac{(\theta - \frac{1}{s_i}\sum_{\ell=1}^{s_i} x^i_\ell)^2}{2}\right),
\end{equation*}
where $\theta\in [-8,8]$ plays the role of a `hyperparameter' that we would like to select. Here, the optimal choice for task $i$ is $\theta = \frac 1 {s_i} \sum_{\ell=1}^{s_i} x^i_\ell$ and hence it is varying together with the underlying mean $\mu^i$ of the sampling distribution. An illustration of this experiment can be found in Figure \ref{fig:toy_illustrate} in Appendix \ref{app:experiment_toy}.
\begin{figure}[t]
    \centering
    \includegraphics[scale=0.25]{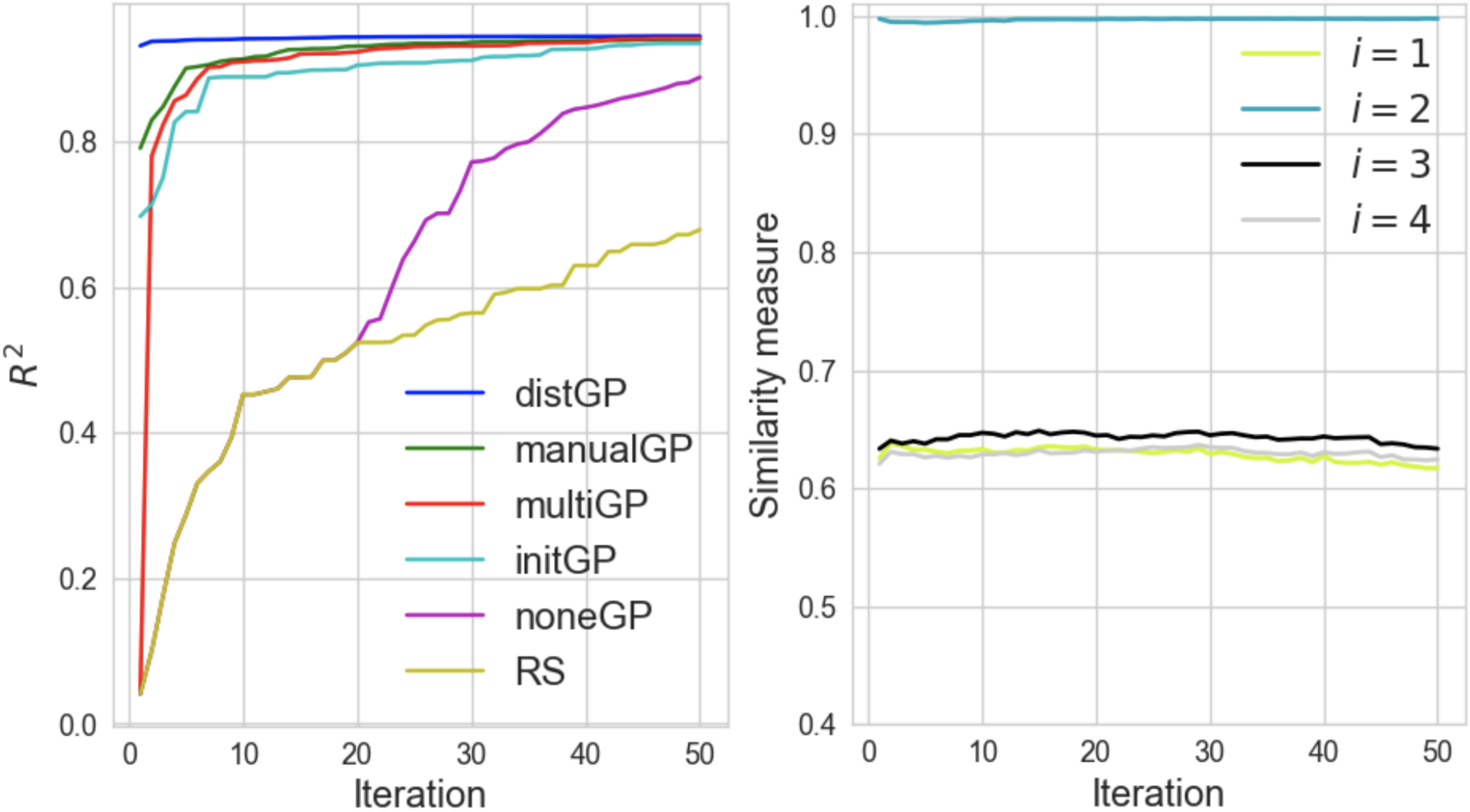}
    \caption{
    Handcrafted meta-features counterexample over $30$ runs, with $50$ iterations \textbf{Left}: Mean of the \textit{maximum observed} $f^{target}$ so far (including any initialisation). \textbf{Right:} Mean of the similarity measure $k_p(\psi(D_i), \psi(D_{\text{target}}))$ for distGP, the target task uses the same generative process as $i=2$.}
    \label{fig:meta}
\end{figure}
We now perform an experiment with $n=15$, and $s_i=500$, for all $i$, and generate $3$ source tasks with $\gamma^i=0$, and $12$ source task with $\gamma^i=4$. In addition, we generate an additional target dataset with $\gamma^{\text{target}}=0$ and let the number of source evaluations per task be $N_i = 30$. 

The results can be found in Figure \ref{fig:toy_sim}. Here, we observe that distBO has correctly learnt to utilise the appropriate source tasks, and is able to few-shot the optimum. This is also evident on the right of Figure \ref{fig:toy_sim}, which shows the similarity measure $k_p(\psi(D_i), \psi(D_{\text{target}})) \in [0,1]$  for distGP. The feature representation has correctly learned to place high similarity on the three source datasets sharing the same $\gamma^i$ and hence having similar values of $\mu^i$, while placing low similarity on the other source datasets. As expected, manualBO also few-shots the optimum here since the mean meta-feature which directly reveals the optimal hyperparameter was explicitly encoded in the hand-crafted ones. initBO starts reasonably well, but converges slowly, since the optimal hyperparameters even in the similar source tasks are not the same as that of the target task. It is also notable that multiBO is unable to few-shot the optimum, as it does not make use of any meta-information, hence needing initialisations from the target task to even begin learning the similarity across tasks. This is especially highlighted in Figure \ref{fig:toy_all_sim} in Appendix \ref{app:experiment_toy}, which shows an incorrect similarity in the first few iterations. Significance is shown in the mean rank graph found in Figure \ref{fig:toy_rank} in Appendix \ref{app:experiment_toy}.\\ \\
\subsection{When handcrafted meta-features fail.}
We now demonstrate an example in which using handcrafted meta-features does not capture any information about the optimal hyperparameters of the target task. Consider the following process for dataset $i$ with $\mathbf x_\ell^i\in\mathbb R^6$ and $y_\ell^i\in\mathbb R$, given by:
\begin{eqnarray}
\left[\mathbf{x}^i_\ell\right]_j & \iid & \mathcal{N}(0, 2^2), \quad j=1,\dots, 6, \nonumber\\
\left[ \mathbf{x}^i_\ell\right]_{i+2} & = & \text{sign}([\mathbf{x}^i_\ell]_1 [\mathbf{x}^i_\ell]_2) \left| [\mathbf{x}^i_\ell]_{i+2} \right|, \label{counter_transform}\\ 
y^i_\ell & = & \log\left(1+ \left( \prod_{j\in \{1, 2, i+2\}} [\mathbf{x}^i_\ell]_j \right)^3\right) + \epsilon^i_\ell.\nonumber
\end{eqnarray}
where $\epsilon^i_\ell \overset{iid}\sim \mathcal{N}(0, 0.5^2)$, with index $i, \ell, j$ denoting task, sample and dimension, respectively: $i=1, \dots, 4$ and $\ell=1,\ldots,s_i$ with sample size $s_i=5000$. Thus across $n=4$ source tasks, we have constructed regression problems, where the dimensions which are relevant (namely $1$, $2$ and $i+2$) are varying. Note that \eqref{counter_transform} introduces a three-variable interaction in the relevant dimensions, but that all dimensions remain pairwise independent and identically distributed.  Thus, while these tasks are inherently different, this difference is invisible by considering marginal distribution of covariates and their pairwise relationships such as covariances. As the handcrafted meta-features for manualBO only consider statistics which process one or two dimensions at the time or landmarkers \citep{pfahringer2000meta}, their corresponding $\psi(D_i)$ are \textit{invariant} to tasks up to sampling variations. For an in-depth discussion, see Appendix \ref{app:counter}. We now generate an additional target dataset, using the same generative process as $i=2$, and let $f$ be the coefficient of determinant ($R^2$) on the test set resulting from an automatic relevance determination (ARD) kernel ridge regression with hyperparameters $\alpha$ and $\sigma_1$, \dots, $\sigma_6$. Here $\alpha$ denotes the regularisation parameter, while $\sigma_j$ denotes the kernel bandwidth for dimension $j$. Setting $N_i=125$, the results can be found in Figure \ref{fig:meta} (GP) and Figure \ref{fig:counter_all} in Appendix \ref{app:counter} (BLR). It is clear that while distBO is able to learn a high similarity to the correct source task (as shown in Figure \ref{fig:meta}), and one-shot the optimum, this is not the case for any of the other baselines (Figure \ref{fig:counter_sim} in Appendix \ref{app:counter}) . In fact, as manualBO's meta-features do not include any useful meta-information, they essentially encode the task index, and hence perform similarly to multiBO. Further, we observe that initBO has slow convergence after warm-starting. This is not surprising as initBO has to  `re-explore' the hyperparameter space as it only uses a subset of previous evaluations. This highlights the importance of using all evaluations from all source tasks, even if they are sub-optimal. In Figure \ref{fig:counter_all} in Appendix \ref{app:counter}, we show significance using a mean rank graph and that the BLR methods performs similarly to their GP counterparts.
\begin{figure}[t]
    \centering
    \includegraphics[scale=0.45]{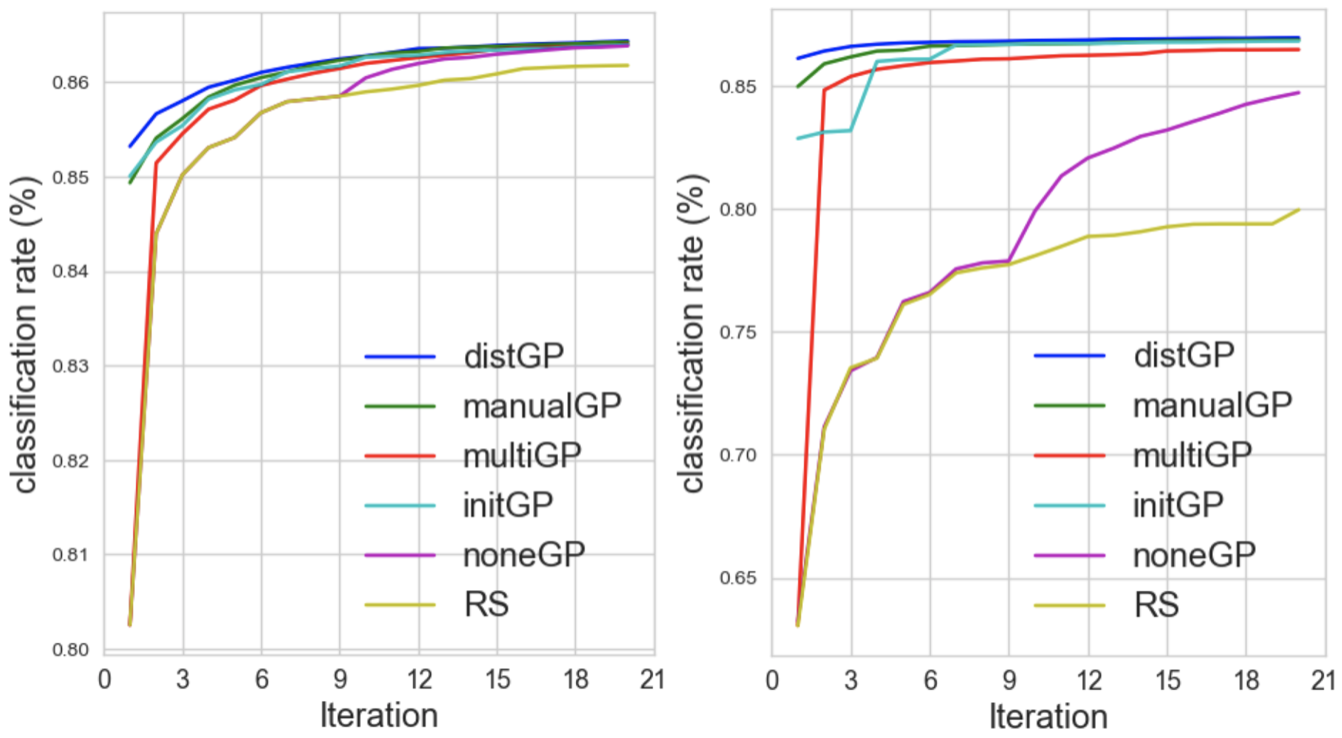}
    \caption{Each evaluation is the \textit{maximum observed} accuracy rate averaged over $140$ runs, with 20 runs on each of the protein as target. \textbf{Left:} Jaccard kernel C-SVM. \textbf{Right:} Random forest}
    \label{fig:protein}
\end{figure}

\subsection{Classification: Protein dataset.} The Protein dataset consists of $7$ different proteins extracted from \cite{gaulton2016chembl}: ADAM17, AKT1, BRAF, COX1, FXA, GR, VEGFR2. Each protein dataset contains $1037-4434$ molecules (data-points $s_i$), where each molecule has binary features $\mathbf{x}^i_\ell \in \mathbb{R}^{166}$ computed using a chemical fingerprint (MACCs Keys\footnote{http://rdkit.org/docs/source/rdkit.Chem.MACCSkeys.html}). The label per molecule is whether the molecule can bind to the protein target $\in \{0, 1\}$. In this experiment, we can treat each protein as a separate classification task. 
We consider two classification methods: Jaccard kernel C-SVM \cite{bouchard2013proof, ralaivola2005graph} (commonly used for binary data, with hyperparameter $C$), and random forest (with hyperparameters $n\_trees$, $max\_depth$, $min\_samples\_split$, $min\_samples\_leaf$), with the corresponding objective $f$ for each given by accuracy rate on the test set. In this experiment, we will designate each protein as the target task, while using the other $n=6$ proteins as source tasks. In particular, we will take $N_i=20$ and hence $N=120$. The results obtained by averaging over different proteins as the target task ($20$ runs per task) are shown in Figure \ref{fig:protein} (with mean rank graphs and BLR version to be found in Figure \ref{fig:protein_svm} and \ref{fig:protein_forest} in Appendix \ref{app:protein}). On this dataset, we observe that distGP outperforms its counterpart baselines and few-shots the optimum for both algorithms. In addition, we can see a slower convergence for the multiGP and initGP, demonstrating the usefulness of meta information in this context.
\section{Conclusion}
We demonstrated that it is possible to borrow strength between multiple hyperparameter learning tasks by making use of the similarity between training datasets used in those tasks. This helped us to develop a method which finds a favourable setting of hyperparameters in only a few evaluations of the target objective. We argue that the model performance should not be treated as a black box function as it corresponds to specific known models and specific datasets and that its careful consideration as a function of all its inputs, and not just of its hyperparameters, can lead to useful algorithms. 

\bibliographystyle{plainnat}
\bibliography{references}
\vfill
\pagebreak
\appendix
\section{Additional details for methodology} 
\label{app:algorithm}
\subsection{Gaussian process (distGP)}
For distGP, we have the following model:
\begin{eqnarray*}
f &\sim & GP(\mu, C)\\
z|\gamma & \overset{i.i.d.}\sim & \mathcal{N}(f(\gamma),\sigma^2)
\end{eqnarray*}
where here $\mu$ is taken to be a constant and $C$ is the corresponding covariance function. In this case, the log marginal likelihood with observations $\Gamma = \left\{\{(\theta^i_k, \mathcal{P}^i_{XY}, s_i), z^i_k\}_{k=1}^{N_i}\right\}_{i=1}^n$, following standard GP literature \citep{rasmussen2004gaussian} is given by:
\begin{equation*}
    \log(p(\mathbf{z}|\Gamma)) = -\frac{1}{2} (\mathbf{z} - \mu)^\top (K + \sigma^2 I)^{-1} (\mathbf{z} - \mu) - \frac{1}{2} \log |K + \sigma^2I| - \frac{N}{2} \log(2\pi)
\end{equation*}
where $\mathbf{z} = [z^1_1, \dots z^n_{N_n}]^\top$, $N = \sum_i N_i$ and $K$ is the kernel matrix, with $K_{ij}= C(\gamma_i, \gamma_j)$. Here $\gamma_i, \gamma_j$ denotes elements of $\Gamma$. In particular, for a new observation $\gamma^\ast$, the predictive posterior distribution $f_{\text{\text{post}}}(\gamma^\ast) \sim \mathcal{N}(\mu_{\text{post}}(\gamma^\ast), \sigma^2_{\text{post}}(\gamma^\ast))$, where:
\begin{eqnarray*}
\mu_{\text{post}}(\gamma^\ast) & = & \mu + K_{\gamma^\ast \Gamma} (K + \sigma^2I)^{-1}(\mathbf{z} - \mu)  \\
\sigma^2_{\text{post}}(\gamma^\ast) & = & K_{\gamma^\ast \gamma^\ast} -  K_{\gamma^\ast \Gamma} (K+\sigma^2 I)^{-1}K_{\gamma^\ast \Gamma}^\top
\end{eqnarray*}
where here $K_{\gamma^\ast \gamma^\ast} = C(\gamma^\ast, \gamma^\ast)$ and $K_{\gamma^\ast \Gamma} = [C(\gamma^\ast, \gamma_1), \dots, C(\gamma^\ast, \gamma_N)]$.

\subsection{Bayesian Linear Regression (distBLR)}
\begin{equation*}
z|\beta  \overset{i.i.d.}\sim \mathcal{N}(\Upsilon \beta,\sigma^2 I )
\quad \quad  \beta \sim  \mathcal{N}(0, \alpha I) 
\end{equation*}
where $\Upsilon = [\upsilon([\theta^1_1, \psi(D_1), s_1]), \dots, \upsilon([\theta^n_{N_n}, \psi(D_n), s_n
])]^\top \in \mathbb{R}^{N\times d}$ and $\alpha > 0$ denotes the prior regularisation. Here $\upsilon$ denotes a feature map of dimension $d$ on concatenated hyperparameters $\theta$, data embedding $\psi(D)$ and sample size $s$. Following \citep{bishop2006, perrone2018scalable}, defining $K_{\text{dim}} = I_{d} + \frac{\alpha}{\sigma^2} \Upsilon^\top \Upsilon$, and $L$ as the cholesky factor of $K_{\text{dim}}$, i.e. $K_{\text{dim}} = LL^\top$, the log marginal likelihood (up to additive constants) with observations $\Gamma = \left\{\{(\theta^i_k, \mathcal{P}^i_{XY}, s_i), z^i_k\}_{k=1}^{N_i}\right\}_{i=1}^n$ is given by:
\begin{equation*}
\log(p(\mathbf{z}|\Gamma)) = \frac{1}{2\sigma^2}(\frac{\alpha}{\sigma^2} ||\mathbf{e}||^2 - ||\mathbf{z}||^2 ) - \sum_{i=1}^d \log(L_{ii}) - \frac{N}{2} \log(\sigma^2)
\end{equation*}
where $\mathbf{e} = L^{-1}\Upsilon^\top \mathbf{z}$. In this case, for a given $\boldsymbol{\upsilon}^\ast \in \mathbb{R}^{d \times 1}$, the transformed feature map of a particular instance of $\gamma^\ast$, the predictive posterior distribution $ \beta^\top \boldsymbol{\upsilon}^\ast = f_{\text{\text{post}}}(\gamma^\ast)  \sim \mathcal{N}(\mu_{\text{post}}(\gamma^\ast), \sigma^2_{\text{post}}(\gamma^\ast))$, where:
\begin{eqnarray*}
\mu_{\text{post}}(\gamma^\ast) & = & \frac{\alpha}{\sigma^2}\mathbf{e}^\top L^{-1} \boldsymbol{\upsilon}^\ast  \\
\sigma^2_{\text{post}}(\gamma^\ast) & = & \alpha ||L^{-1}\boldsymbol{\upsilon}^\ast ||^2
\end{eqnarray*}
It is noted that the computational complexity here scales linearly in the number of observations $N$ and cubically in $d$. 
\subsection{Warm-starting, acquisition functions and multi-task extension}
\label{app:warm-start}
The lower confidence bound (LCB) \citep{srinivas2009gaussian} is defined as follows:
\begin{equation*}
    \alpha_{\text{LCB}}(\gamma;f_{\text{post}}) = \mu_{\text{post}}(\gamma) - \kappa * \sigma_{\text{post}}(\gamma)
\end{equation*}
where $\kappa$ denotes the level of exploration, and for experiments we set $\kappa = 2.58$, as we would like to exploit the information from other tasks on our first iteration. It should be noted that this is not the upper confidence bound commonly used, as we would like to penalise uncertainty on the first iteration.\\ \\
The expected improvement (EI) \citep{movckus1975bayesian} is defined as follows:
\begin{eqnarray*}
    g(\gamma) &=& (\mu_{\text{post}}(\gamma) - z_{\text{max}} - \xi)/\sigma_{\text{post}}(\gamma)\\
    \alpha_{\text{EI}}(\gamma;f_{\text{post}}) & = & \sigma_{\text{post}}(\gamma)(g(\gamma) \Phi_{\text{cdf}}(g(\gamma)) + \mathcal{N}(g(\gamma); 0,1)
\end{eqnarray*}
where here $z_{\text{max}}$ refers to the maximum observed $z$ for our \textit{target task}, while $\Phi_{\text{cdf}}$ and $\mathcal{N}(g(\gamma); 0,1)$ refers to the CDF and pdf of a standard Normal distribution. For experiments, we set the exploration parameter to be $\xi = 0.01$. It should be noted in the case, where the $\alpha_{\text{EI}} = 0$ (or numerically close to $0$) for all attempted locations, we will use the upper confidence bound (with $\kappa=2.58$) \citep{srinivas2009gaussian} instead. To maximise the acquisition function, we first randomly select $300,000$ hyperparameters for evaluation (computationally cheap), to find the top $10$ optimum. Initialising from these top $10$ hyperparameters, a L-BFGS-B algorithm (computationally expensive) is used to maximise the acquisition function, to select the next hyperparameter for evaluation.

\paragraph{Warm-starting} Instead of using the LCB acquisition function (for the first evaluation), an alternative approach is to warm-start \citep{gomes2012combining,reif2012meta,feurer2015initializing} based on \textit{learnt} similarities with previous source tasks. For the GP case, we will optimise the marginal likelihood based on all observations from the source tasks, learning the task similarity function $k_p([\psi(D_i), s_i],[\psi(D_j), s_j])$. As the output domain of $k_p$ lies in $[0,1]$, we can compute the top $M$ source tasks most similar with our target task. Given this selection, we can extract the best $m$ previous best hyperparameters from each of these source tasks, enabling $Mm$ hyperparameters as warm-start initialisations for our algorithm. For the BLR case, as a joint space over $\theta$, $\psi(D)$ and $s$ is considered, a direct task similarity function is no longer available. Instead we opt for a different approach and extract $m$ previous best hyperparameters from all source tasks, and consider only these hyperparameters for the maximisation of the LCB/EI acquisition function. In practice, we recommend to warm-start with as few evaluations as possible, as:
\begin{itemize}
    \item Source tasks can be dissimilar to our target task. 
    \item Warm-start hyperparameters may be similar to each other, and hence costly evaluations are either wasted or inefficient. 
    \item More evaluations are needed before the proposed algorithm can begin to utilise all seen evaluations to explore/exploit for our target task. 
\end{itemize}
\section{Baselines}
\label{app:baselines}
\subsection{manualBO}
\label{app:manual}
Instead of constructing $\psi(D)$, as described in section \ref{sec:method}, we can select $\psi(D)$ to be a selection of handcrafted meta-features. Here, we provide the set of meta-features we used for experiments. It should be noted that features of $X^i = \{\mathbf{x}^i_\ell\}_{\ell=1}^{s_i}$ is standardised to have mean $0$ and variance $1$ individually (except for the unsupervised toy example case, in which we encode the mean meta-feature explicitly), while $y^i_\ell$ is normalised to be in $[0, 1]$ for regression. To ensure fair relative comparisons, meta-features are normalised to be in $[0,1]$ across all tasks \citep{bardenet2013collaborative}. We do not include sample size $s_i$, as these are already encoded separately. 
\paragraph{General meta-features}
\begin{itemize}
    \item Skewness, kurtosis \citep{normalise}: these are calculated on each feature of the dataset $X^i$, before the minimum, maximum, mean and standard deviation of the computed quantities is extracted across the features.
    \item Correlation, covariance \citep{normalise}: these are calculated on every pair of features of $X^i$, before the minimum, maximum, mean and standard deviation of the computed quantities is extracted across each pair of features.
    \item PCA skewness, kurtosis \citep{feurer2014using}: principal component analysis (PCA) is performed on $X^i$, and $X^i$ is projected onto the first principal component. The corresponding skewness and kurtosis is computed.
    \item Intrinsic dimensionality \citep{bardenet2013collaborative}: number of principal components to explain $95\%$ of variance.
\end{itemize}
\paragraph{Classification specific meta-features}
\begin{itemize}
    \item Class ratios, entropy \citep{normalise}: empirical class distribution and its corresponding entropy.
    \item Classification landmarkers \citep{pfahringer2000meta}: 1-nearest-neighbour classifier, linear discriminant analysis, naive Bayes and decision tree classifier.
\end{itemize}
\paragraph{Regression specific meta-features}
\begin{itemize}
    \item Mean, standard deviation, skewness, kurtosis of the labels  $\{y^i_\ell\}_{\ell=1}^{s_i}$ \citep{normalise}.
    \item Regression landmarkers \citep{pfahringer2000meta}: 1-nearest-neighbour regressor, linear regression and decision tree regressor.
\end{itemize}
The landmarkers are scalable algorithms that are cheap to run, and provide us various characteristic of the machine learning task. The corresponding meta-feature from these landmarkers is the accuracy on an independent set of data (a train-test split is done on $X^i$, the training data). In experiments, we use the default settings in \textit{sklearn} \citep{scikit-learn} for these algorithms. For additional details on their formulation and rationale, please refer to \citep[Ch.2]{automl_book}.
\subsection{multiBO}
Instead of using meta-features, we may wish to simply encode the task index, and learn task similarities based on only $\left\{\{\theta^i_k, z^i_k\}_{k=1}^{N_i}\right\}_{i=1}^n$. It should be noted that in both these cases, we do not encode any sample size or class ratio information and initial evaluations from the target task is required. 
\paragraph{multiGP}
For the GP case, we will follow \citep{swersky2013multi}, who considers a multi-task GP for Bayesian optimisation. Instead of using the kernel $k_p$ on meta-features, we will now replace it by a kernel on tasks $k_t$. Given the $n+1$ total number of tasks (including the target task), the task similarity matrix is given by $S_t = L_t L_t^T\in \mathbb{R}^{n+1 \times n+1}$, where $L_t$ is a learnt cholesky factor. 
Expanding $S_t$ into the appropriate sized kernel $K_t \in \mathbb{R}^{N\times N}$ (as we have repeated observations from the same task), using the marginal likelihood, we can learn the lower triangular elements of $L_t$. Similar to \citep{swersky2013multi}, we assume positive correlation amongst tasks and restrict positivity in the elements of the cholesky factor.
\paragraph{multiBLR}
For the BLR case, we will follow \citep{perrone2018scalable} and consider a one-hot encoding for $\psi(D_i)$. This representation essentially identifies a separate encoding for every task, and similarity between tasks (and hyperparameters) is captured through the transformation $\upsilon$ (without sample size $s_i$), which we learn using the marginal likelihood.
\subsection{initBO}
For this baseline, we will employ the handcrafted meta-features as described in Appendix \ref{app:manual} to warm-start Bayesian optimisation, using a GP or BLR. In particular, we first define the number of evaluations $m$ per task and the number of tasks $M$ we wish to warm-start with (i.e. $Mm$ number of warm-start hyperparameters). To define a similarity function, for a fair comparison with existing literature, we will use the $\ell_2$ norm \citep{feurer2015initializing} between the datasets' meta-features:
\begin{equation*}k(D_i,D_j) = -|| \ [\psi(D_i), s_i] - [\psi(D_j), s_j] \ ||_2 \end{equation*}
where here $k$ is a similarity function, and $\psi(D_i)$ is the handcrafted meta-features representation for task $i$. It should also be noted that as meta-features are individually normalised to be in $[0,1]$, no particular meta-feature is emphasised in this distance measure. To obtain the warm-start $\theta$s, we compute $k(D_{\text{target}}, D_j)$ for all $j=1, \dots, n$ and extract the $M$ tasks with highest similarity. Given these $M$ tasks, we extract the $m$ best performing hyperparameters from each of these task to obtain $Mm$ warm-start hyperparameters. These hyperparameters will then be used for warm-starting noneGP or noneBLR (instead of random evaluations).

\section{Experiments}
\label{app:experiment}
With the exception of the hyperparameter in the unsupervised toy and the protein random forest example, all other hyperparameters are optimised in the log-scale. In addition, we standardise hyperparameters to have mean 0 and variance 1, when passing them to the GP and BLR, to ensure parameters initialisation are well-defined. Here we provide additional details for our experiments in section \ref{sec:experiments}.
\subsection{Comparison between joint and concatenation embeddings for regression}
\label{app:concat_embed}
Here we display additional graphs comparing the embedding of the joint distribution versus the embedding of the conditional distribution and marginal distribution before concatenation. We denote these correspondingly by distGP-joint, distBLR-joint and distGP-concat, distGP-concat. Overall, we observe that their performance is similar.
\begin{figure}[ht!]
    \centering
    \includegraphics[scale=0.5]{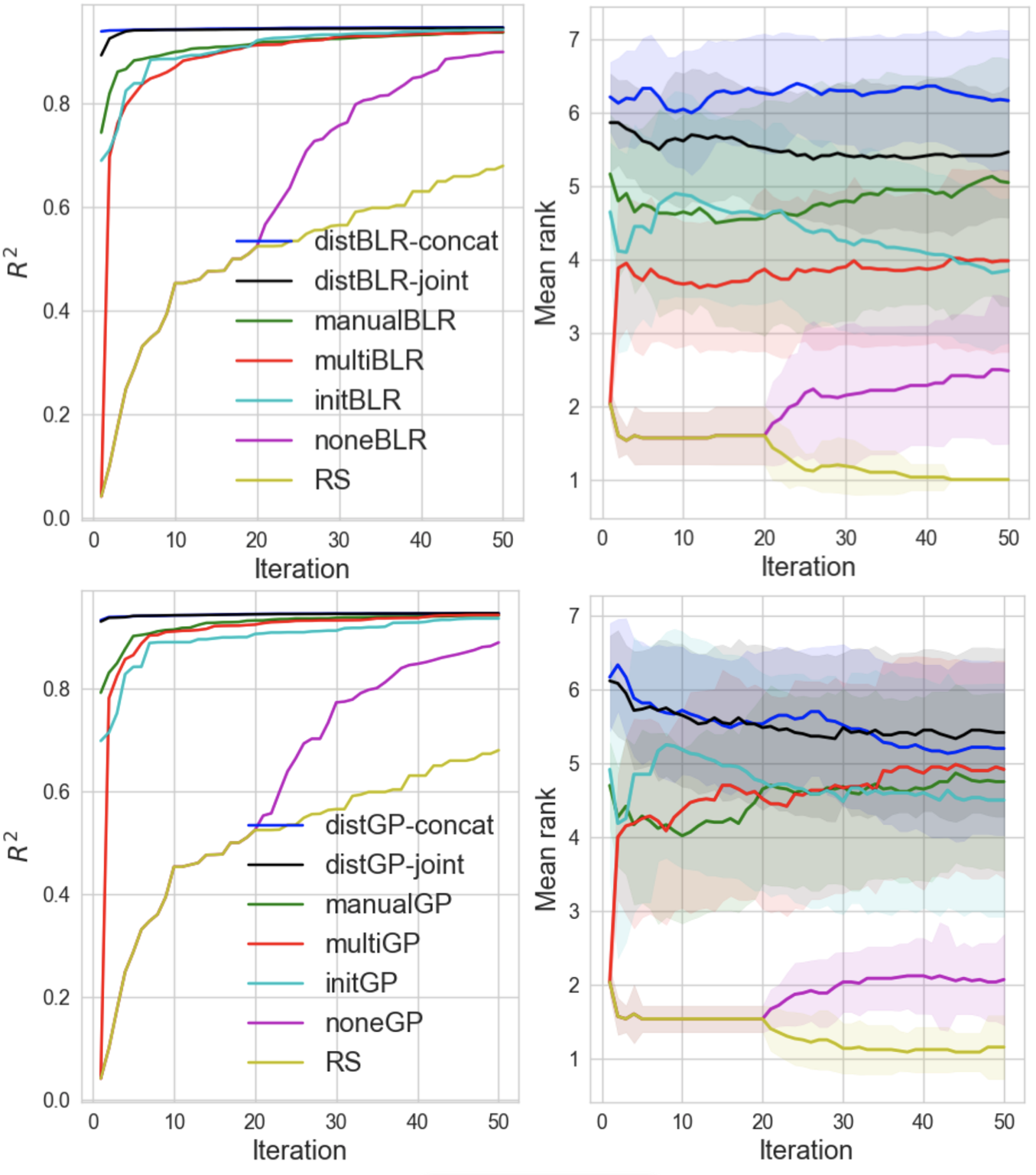}
    \caption{Manual meta-features counterexample with $50$ iterations (including any initialisation). Here, BLR methods are displayed on the top, while GP methods are displayed on the bottom. Each evaluation here is averaged over $30$ runs. \textbf{Left:} \textit{Maximum observed} $R^2$. \textbf{Right:} Mean rank (with respect to each run) of the different methodologies, with $\pm1$ sample standard deviation.}
    \label{fig:concat_joint_counter}
\end{figure}

\begin{figure}[ht!]
    \centering
    \includegraphics[scale=0.42]{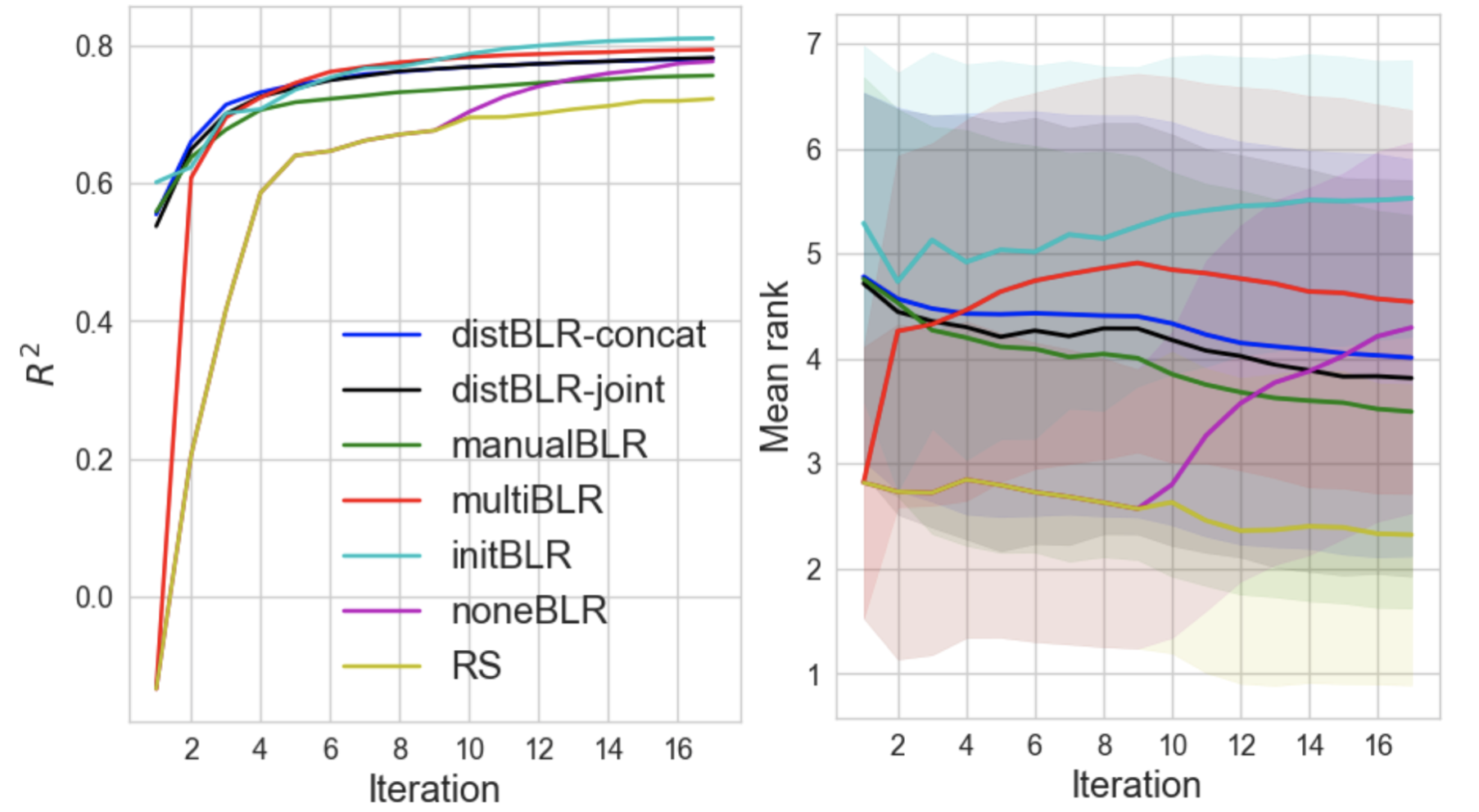}
    \caption{Parkinson's experiment with $17$ iterations (including any initialisation). Each evaluation here is averaged over $420$ runs, with each of the $42$ patient set as the target task (repeated for 10 runs) \textbf{Left:} \textit{Maximum observed} $R^2$. \textbf{Right:} Mean rank (with respect to each run) of the different methodologies, with $\pm1$ sample standard deviation.}
    \label{fig:park_vary}
\end{figure}
\FloatBarrier
\subsection{Unsupervised toy example}
\label{app:experiment_toy}
\begin{figure}[ht!]
    \centering
    \includegraphics[scale=0.5]{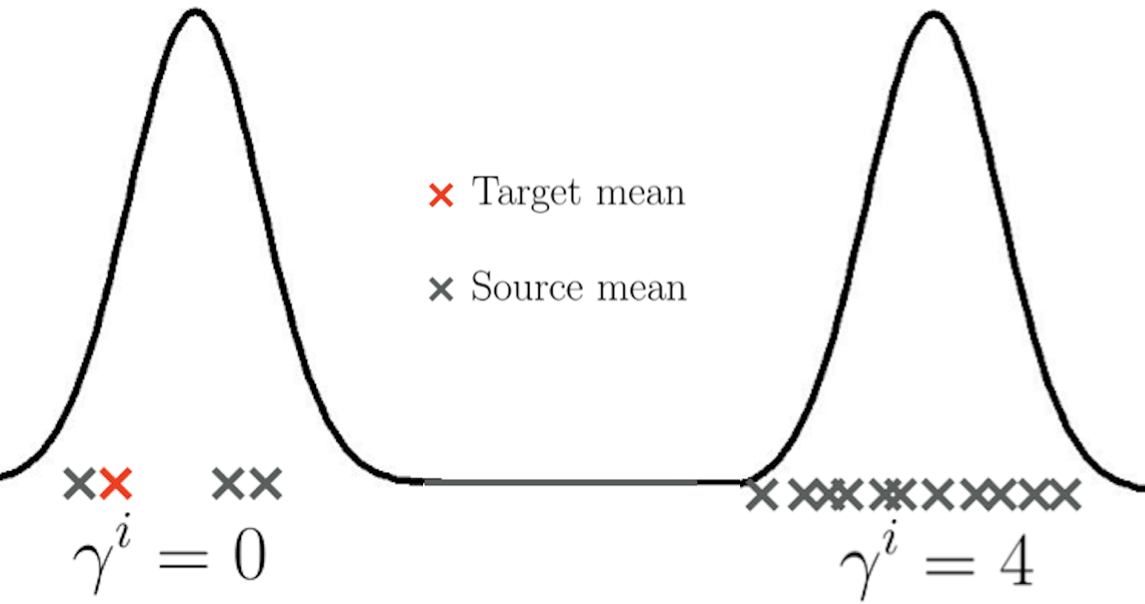}
    \caption{Illustration of unsupervised toy example.}
    \label{fig:toy_illustrate}
\end{figure}

Hyperparameters: $\theta \in [-8, 8]$ \\
Source task's random and BO iterations: $10, 20$ \\
Target task's noneBO random and BO iterations: $5, 10$ \\
An illustration of this toy example can be seen in figure \ref{fig:toy_illustrate}. \\ 

\begin{figure}[ht!]
    \centering
    \includegraphics[scale=0.37]{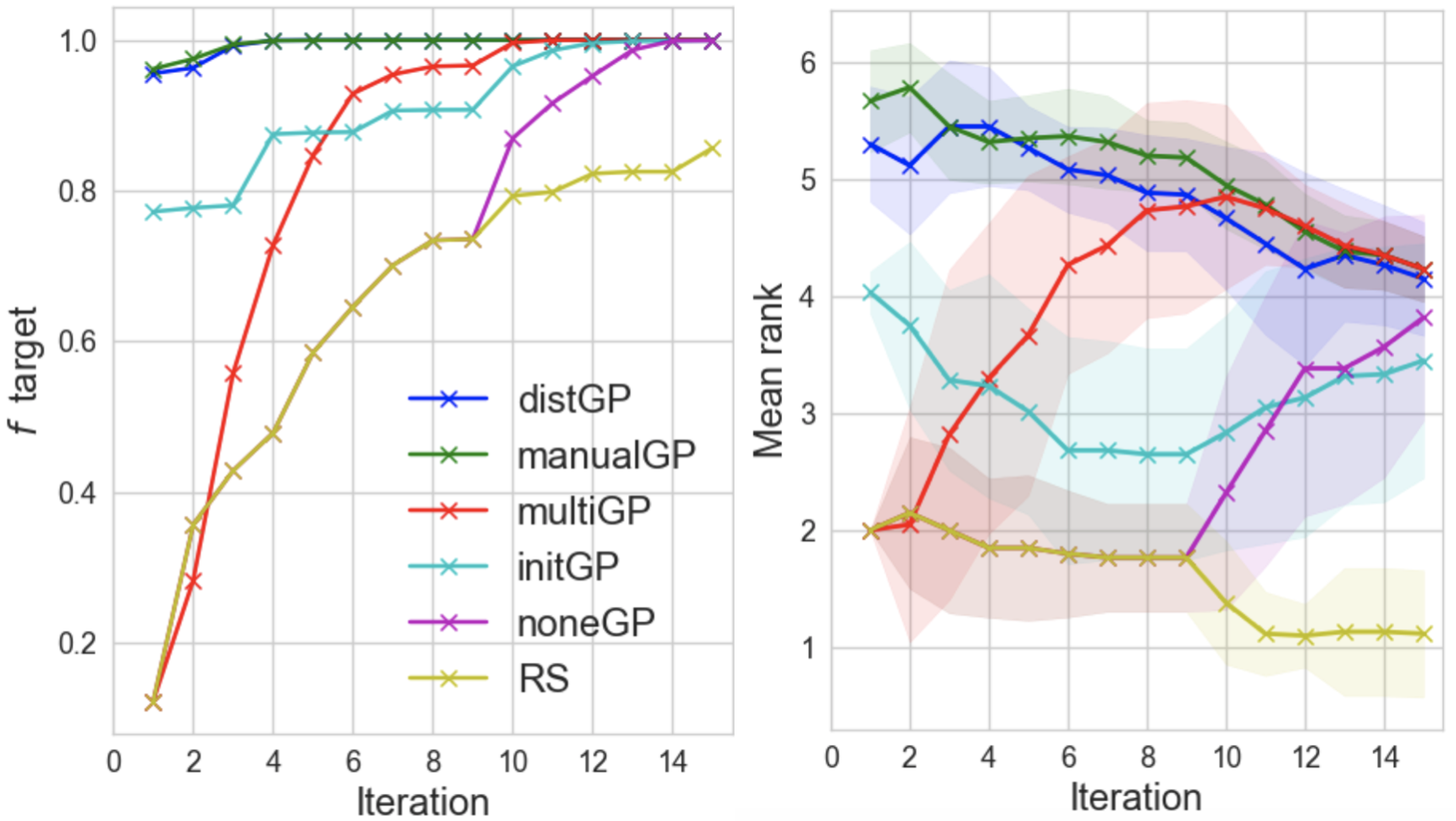}
    \caption{Unsupervised toy task with $15$ iterations (including any initialisation). Each evaluation here is averaged over $30$ runs. \textbf{Left:} \textit{Maximum observed} $f^{target}$. \textbf{Right:} Mean rank (with respect to each run) of the different methodologies, with $\pm1$ sample standard deviation.}
    \label{fig:toy_rank}
\end{figure}

\begin{figure}[ht!]
    \centering
    \includegraphics[scale=0.28]{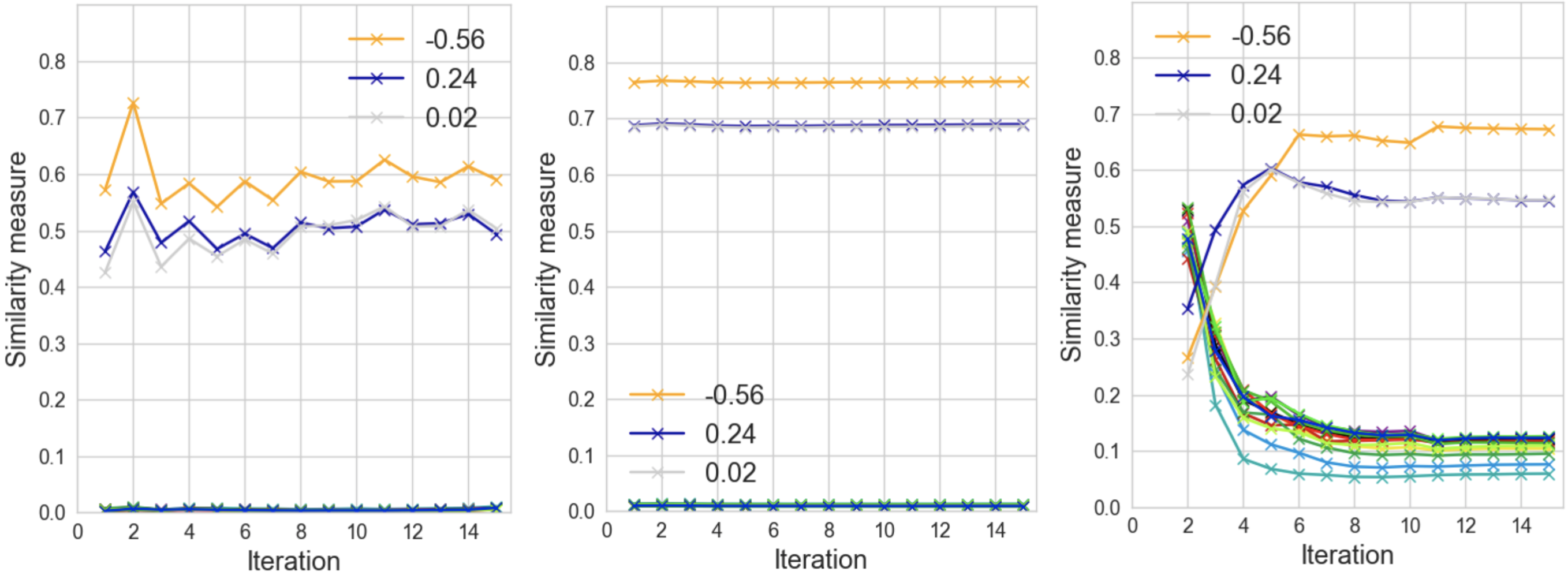}
    \caption{Mean of the similarity measure $k_p(\psi(D_i), \psi(D_{\text{target}}))$ over $30$ runs versus number of iterations for the unsupervised toy task. For clarity purposes, the legend \textit{only} shows the $\mu^i$ for the $3$ source tasks that are similar to the target task with $\mu^i=-0.25$. It is noted the rest of the source task have $\mu^i\approx 4$. \textbf{Left:} distGP \textbf{Middle:} manualGP \textbf{Right:} multiGP}
    \label{fig:toy_all_sim}
\end{figure}
\FloatBarrier
\subsection{Regression: handcrafted meta-features counterexample}
\label{app:counter}
Hyperparameters: $\alpha \in [10.0^{-8}, 0.1], \sigma_j \in [2.0^{-7}, 2.0^5]$ \\
Source task's random and BO iterations: $50, 75$ \\
Target task's noneBO random and BO iterations: $20, 30$

For task $i=1,\dots 4$, we have the process:
\begin{eqnarray*}
\left[\mathbf{x}^i_\ell\right]_j & \sim & \mathcal{N}(0, 2^2) \quad j=1,\dots, 6 \\
\left[ \mathbf{x}^i_\ell\right]_{i+2} & = & \text{sign}([\mathbf{x}^i_\ell]_1 [\mathbf{x}^i_\ell]_2) \left| [\mathbf{x}^i_\ell]_{i+2} \right| \\ 
y^i_\ell & = & \log\left(1+ \left( \prod_{j\in \{1, 2, i+2\}} [\mathbf{x}^i_\ell]_j \right)^3\right) + \epsilon^i_\ell
\end{eqnarray*}
where $\epsilon^i_\ell \overset{iid}\sim \mathcal{N}(0, 0.5^2)$, with index $i, \ell, j$ denoting task, sample and dimension. For each task $i$, the dimension of importance is $1,2$ and $i+2$, while the rest is nuisance variables. We now demonstrate that the handcrafted meta-features for regression in Appendix \ref{app:manual} do not differ across the tasks (when noise is not considered). Firstly, it is noted that $\left[ \mathbf{x}^i_\ell \right ]_{i+2} \sim \mathcal{N}(0, 2^2)$ even after alteration. This then implies that meta-features measuring skewness and kurtosis per dimension does not change across tasks. Similarly, any PCA meta-features will remain the same, as variances remains the same in all directions. Further, as $\left[ \mathbf{x}^i_\ell \right ]_{i+2}$ remains independent to $\left[ \mathbf{x}^i_\ell \right ]_{j}$ for $j \neq k$, meta-features based on correlation and covariance will remain to be $0$ for all pairs of features. Lastly, for regression landmarkers and labels, as these are not perturbed by permutation of the features of the dataset, the regression specific meta-features also remains the same. Together, this implies that the handcraft meta-features are unable to distinguish which source task is similar to the target task (with the same process as $i=2$). However, as we have additional noise samples for each task, the computed representation $\psi(D_i)$ still differs amongst all the tasks, hence the specific task can still be recognised. 

\begin{figure}[ht!]
    \centering
    \includegraphics[scale=0.57]{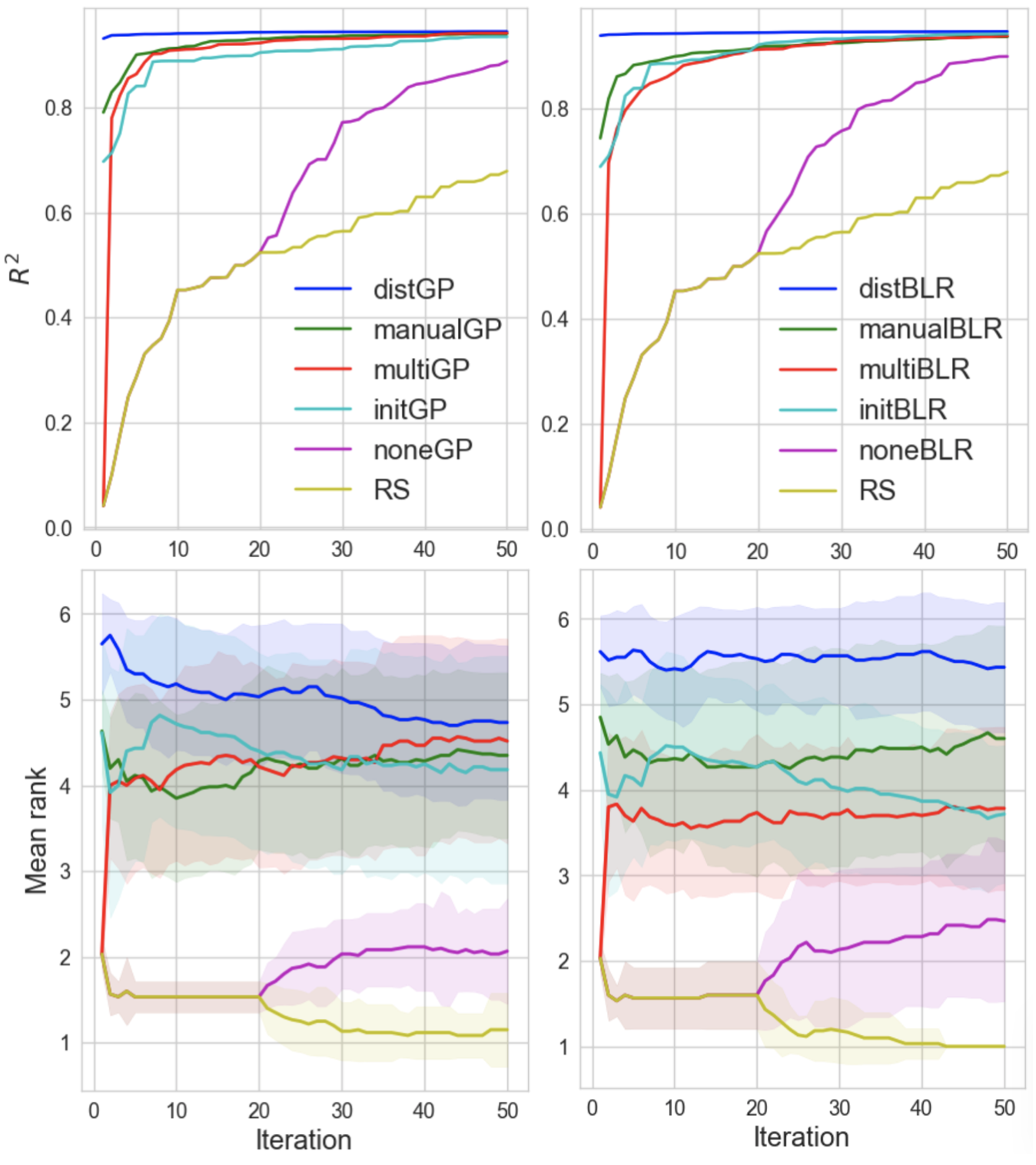}
    \caption{Manual meta-features counterexample with $50$ iterations (including any initialisation). Here, GP methods are displayed on the left, while BLR methods are displayed on the right. Each evaluation here is averaged over $30$ runs. \textbf{Top row:} \textit{Maximum observed} $R^2$. \textbf{Bottom row:} Mean rank (with respect to each run) of the different methodologies, with $\pm1$ sample standard deviation.}
    \label{fig:counter_all}
\end{figure}

\begin{figure}[ht!]
    \centering
    \includegraphics[scale=0.33]{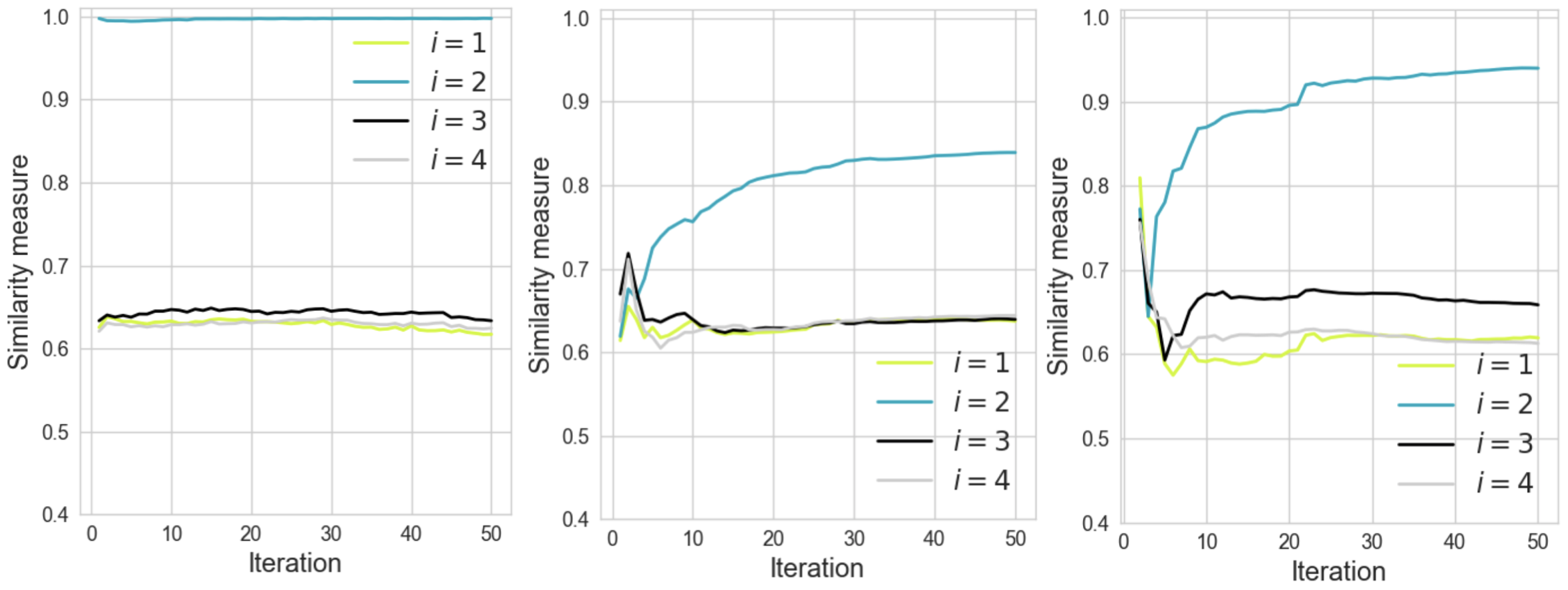}
    \caption{Mean of the similarity measure $k_p(\psi(D_i), \psi(D_{\text{target}}))$ over $30$ runs versus number of iterations for the manuak meta-features counterexample. The target task uses the same generative process as $i= 2$. \textbf{Left:} distGP \textbf{Middle:} manualGP \textbf{Right:} multiGP}
    \label{fig:counter_sim}
\end{figure}
\FloatBarrier
\subsection{Classification: similar and not similar source tasks}
\label{app:unseen_task}
Hyperparameters: $C \in [2.0^{-7}, 2.0^{10}], \sigma_j \in [2.0^{-3}, 2.0^5]$ \\
Source task's random and BO iterations: $75, 75$ \\
Target task's noneBO random and BO iterations: $25, 75$
\begin{figure}[ht!]
    \centering
    \includegraphics[scale=0.29]{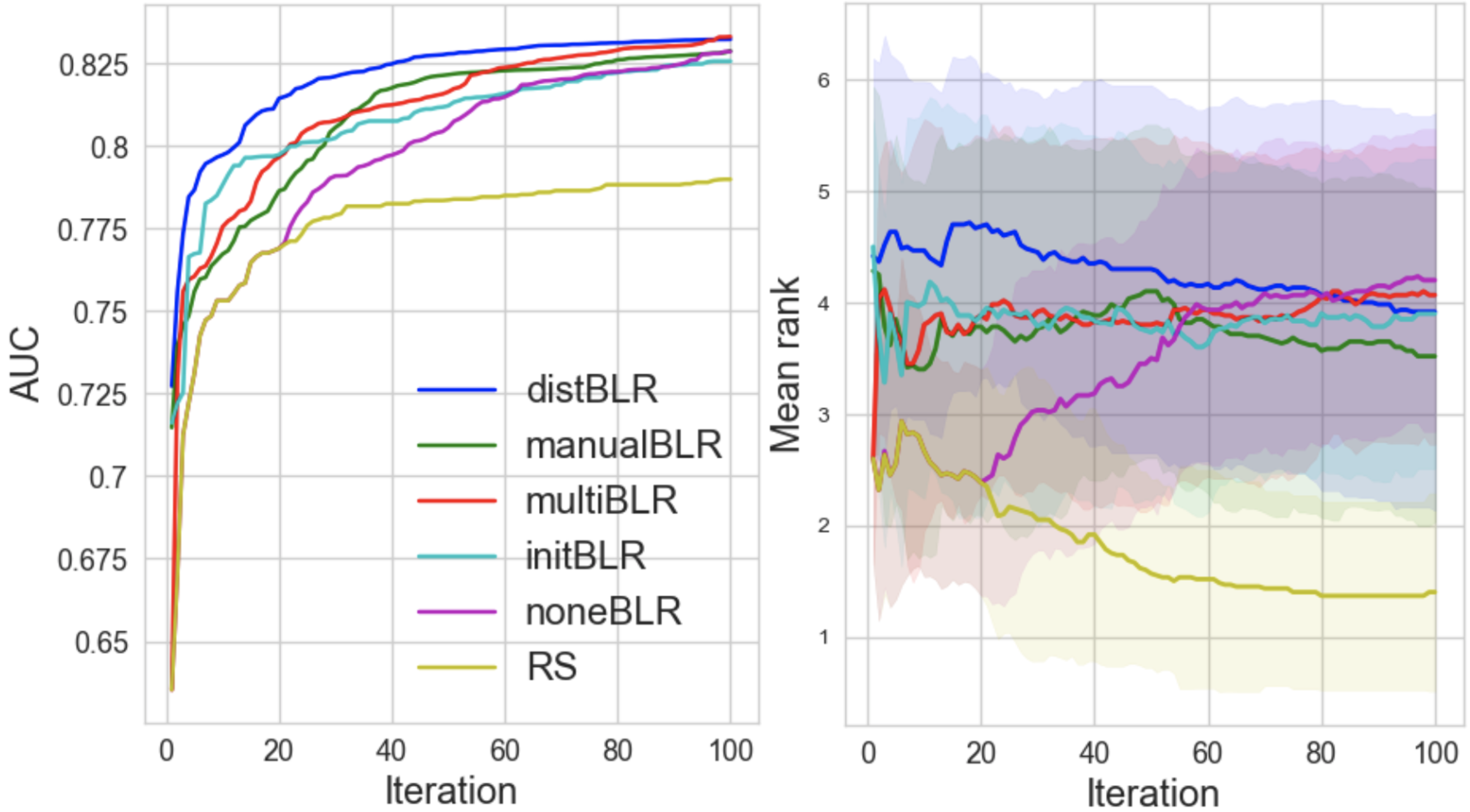}
    \caption{Classification task experiment A with $100$ iterations (including any initialisation). Here, the target task is similar to one of the source task. Each evaluation here is averaged over $30$ runs. \textbf{Left:} \textit{Maximum observed} AUC. \textbf{Right:} Mean rank (with respect to each run) of the different methodologies, with $\pm1$ sample standard deviation.}
    \label{fig:class_bw_0}
\end{figure}

\begin{figure}[ht!]
    \centering
    \includegraphics[scale=0.29]{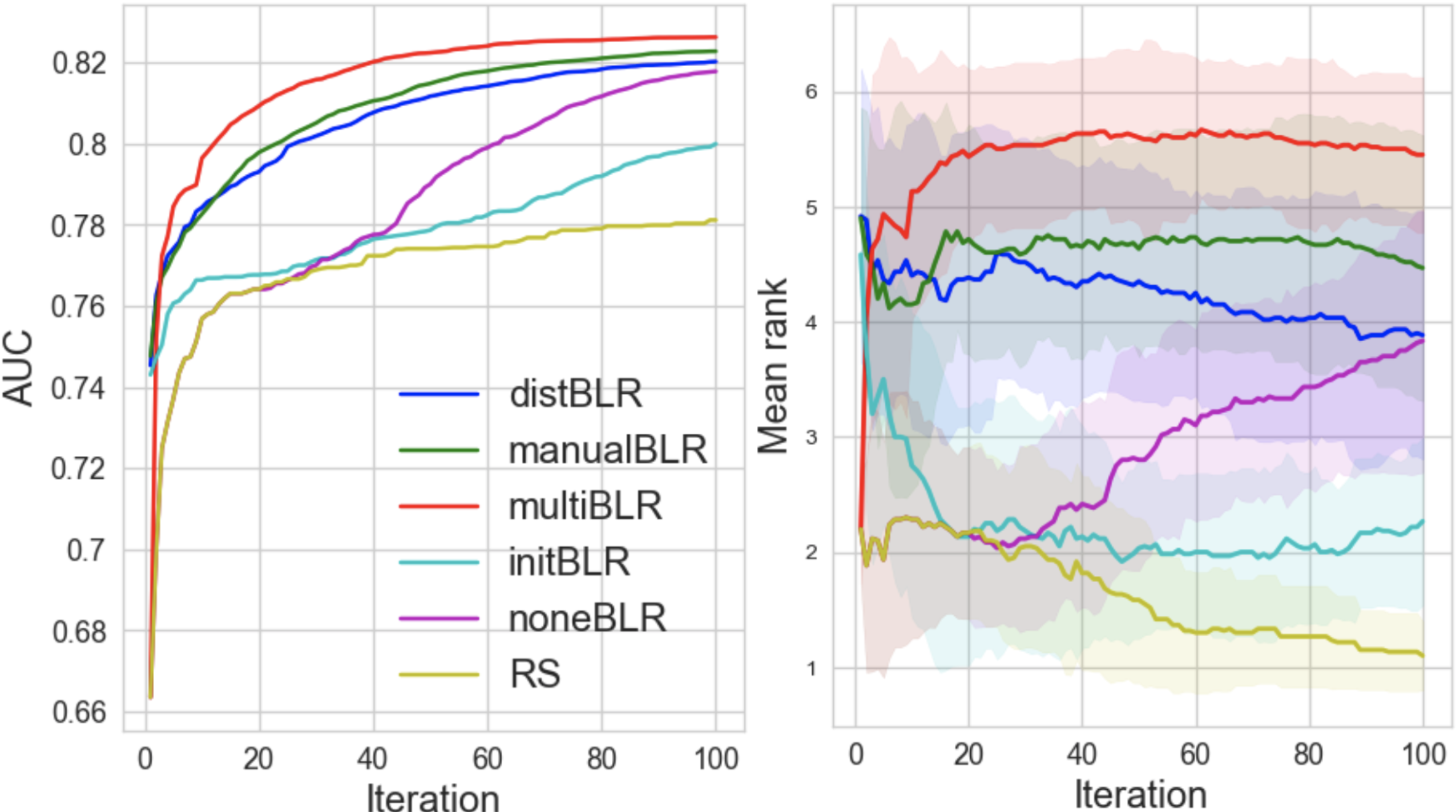}
    \caption{Classification task experiment B with $100$ iterations (including any initialisation). Here the target task is \textit{different} to all the source task. Each evaluation here is averaged over $30$ runs. \textbf{Left:} \textit{Maximum observed} AUC. \textbf{Right:} Mean rank (with respect to each run) of the different methodologies, with $\pm1$ sample standard deviation.}
    \label{fig:class_bw_-1}
\end{figure}
We now demonstrate a classification example, where we contrast the case where some of the source tasks is similar to the target tasks against the case where no such source task exists to illustrate that encoding meta-information need not always be beneficial. Here, we let the number of source tasks $n=10$, $s_i=5000$ and $f$ to be the AUC on the test set for  ARD kernel logistic regression, with hyperparameters $C$ and $\sigma_1$, \dots, $\sigma_6$. Similar to before, $C$ denotes regularisation and $\sigma_j$ denotes the kernel bandwidth for dimension $j$. To generate $D_i$, we take $\mathbf{x}^i_\ell \sim \mathcal{N}(\mathbf{0}, I_6)$, and obtain $y^i_\ell$ conditionally on $\mathbf{x}^i_\ell$ by sampling from a kernel logistic regression model (ARD kernel with Random Fourier features \citep{rahimi2008random} approximation) where each task has different ``true'' bandwidth parameters (also different across dimensions). 

To be more precise, to generate $\{\mathbf{x}^i_\ell, y^i_\ell\}_{\ell=1}^{s_i}$ for this experiment, we first simulate $\mathbf{x}^i_\ell \sim \mathcal{N}(\mathbf{0}, I_6)$. Then in order to sample from the model of an ARD kernel logistic regression, we define an underlying true bandwidth $\tilde{\boldsymbol{\sigma}}^i = [\tilde{\sigma}^i_1, \dots, \tilde{\sigma}^i_6]$ and use random Fourier features (RFF) \citep{rahimi2008random} to approximate an ARD kernel (with $D=200$ frequencies) as follows:
\begin{equation*}
    \boldsymbol{\varphi}^i_\ell = \sqrt{2/D}\cos(\mathbf{U}\tilde{\mathbf{x}}^i_\ell + \mathbf{b}) \quad \quad \mathbf{U} \in \mathbb{R}^{D \times 6}, \mathbf{b}\in \mathbb{R}^{D}
\end{equation*}
where $\tilde{\mathbf{x}}^i_\ell = \mathbf{x}^i_\ell / \tilde{\boldsymbol{\sigma}}^i$ denotes element-wise division by the bandwidths in respective dimensions and $\mathbf{U}_{mn} \overset{i.i.d.}\sim \mathcal{N}(0,1)$ and $\mathbf{b}_m \overset{i.i.d.}\sim \text{Unif}([0, 2\pi])$. Letting $\boldsymbol{\Phi}^i = [\boldsymbol{\varphi}^i_1, \dots \boldsymbol{\varphi}^i_{s_i}]^\top$, we let $\tilde{\mathbf{g}}^i = \boldsymbol{\Phi}^i \boldsymbol \beta^i$, where $\boldsymbol\beta^i \sim \mathcal{N}(0, I_{D})$. We then normalise $\tilde{\mathbf{g}}^i$ to be in the range $[-6, 6]$ and then transform it through the logistic link:
\begin{equation*}
p^i_\ell = \dfrac{1}{1 + \exp(-\tilde{g}^i_\ell)}
\end{equation*}
obtaining $p^i_\ell = P(y^i_\ell=1 | x^i_\ell)$, using which we can draw a binary output $y^i_\ell\sim\text{Bernoulli}(p^i_\ell)$. For the source tasks, we will randomly select $ \tilde{\sigma}^i_j \in \{0.5, 1.0, 2.0, 4.0, 8.0, 16.0\}$ with replacement across all $j$, so that different dimensions are of different relative importance across different tasks. For experiment A, we will select its underlying bandwidths to be the same as one of that in the source task. For experiment B, to ensure that our target task has different optimal hyperparameters to the source tasks, we will let $\tilde{\sigma}^i_j = 1.5 $ for all $j$.

Note that all tasks have the same marginal distribution of covariates and that there is a high variation in conditional distributions: they differ not only in terms of kernel bandwidths but also in terms of coefficients in their respective regression functions. 
To generate a task dataset, we use the same process, and run 2 experiments: (A) use the same set of bandwidths as one of the source tasks but a different regression function, and (B) use a set of bandwidths unseen in any of the source tasks (and a different regression function). 
We take $N_i=150$ and since the total number of evaluations is $N=1500$, we focus our attention on BLR, which have $O(N)$ linear computational complexity. The results for the two experiments are shown in Figure \ref{fig:class_bw_0} and \ref{fig:class_bw_-1}. We see that distBLR leverages the presence of a similar task among the sources and learns a representation of the dataset which helps guide hyperparameter selection to the optimum faster than other methods. We note that manualBLR converges much slower, given that the optimal hyperparameters depend on the data in a complex way which is difficult to extract from handcrafted meta-features. We also note that initBLR performs poorly despite the presence of a source task with the \emph{same ``true'' bandwidths}: often, the meta-features are not powerful enough to recognize which task is the most similar in order to initialise appropriately. On the other hand, in the case B, no similar source exists implying that the joint BLR model in distBLR needs to extrapolate to the far away region in the space of joint distributions of training data. As expected, meta-information in this example is not as helpful as in the case A and the method that ignores it, multiBLR, in fact performs best. However, albeit worse performing, note that distBLR and manualBLR were still able to revert to the behaviour akin to multiBLR and achieve a faster convergence compared to their non-transfer counterparts and initBLR which essentially has to re-explore the hyperparameter space from scratch.  
\FloatBarrier
\subsection{Regression: Parkinson's dataset}
Hyperparameters: $\alpha \in [10.0^{-10}, 0.1], \sigma_j \in [2.0^{-7}, 2.0^5]$ \\
Source task's random and BO iterations: $10, 20$ \\
Target task's noneBO random and BO iterations: $9, 8$

\label{app:parkinson}
\begin{figure}[ht!]
    \centering
    \includegraphics[scale=0.43]{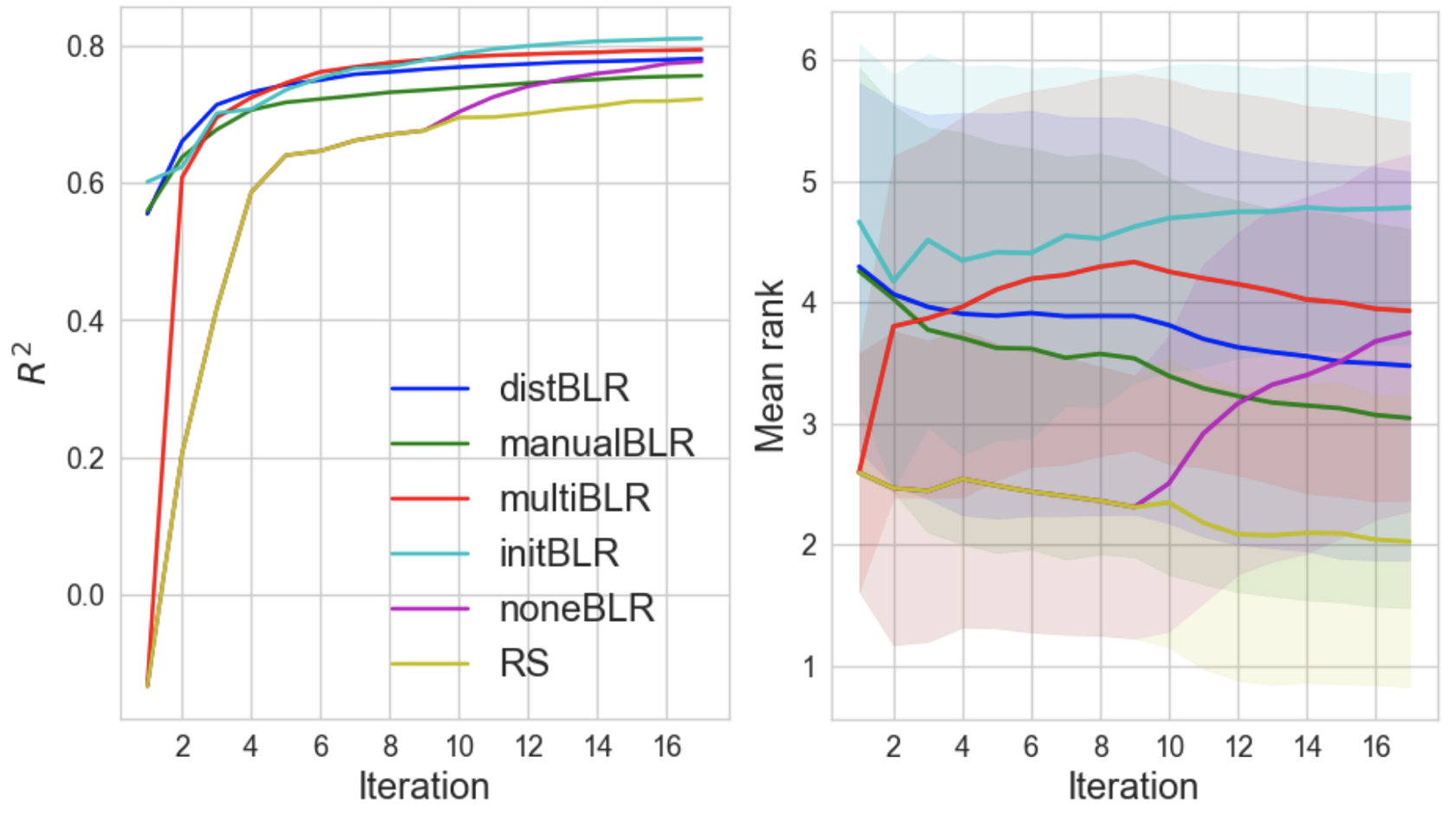}
    \caption{Parkinson's experiment with $17$ iterations (including any initialisation). Each evaluation here is averaged over $420$ runs, with each of the $42$ patient set as the target task (repeated for 10 runs) \textbf{Left:} \textit{Maximum observed} $R^2$. \textbf{Right:} Mean rank (with respect to each run) of the different methodologies, with $\pm1$ sample standard deviation.}
    \label{fig:patient}
\end{figure}
The Parkinson’s disease telemonitoring dataset\footnote{http://archive.ics.uci.edu/ml/datasets/Parkinsons+Telemonitoring} consists of voice measurements using a telemonitoring device for $42$ patients with Parkinson disease (approximately $150$ recordings $\in \mathbb{R}^{17}$ each). The label is the clinician’s Parkinson disease symptom score for \textit{each recording}. Following a setup similar to \cite{blanchard2017domain}, we can treat each patient as a separate regression task. In this experiment, in order to allow for comprehensive benchmark comparisons, we consider $f$ which is not prohibitively expensive (hence the problem does not necessarily benefit computationally from Bayesian optimisation). Namely, we employ RBF kernel ridge regression (with hyperparameters $\alpha$, $\gamma$), with $f$ as the coefficient of determination ($R^2$). In this experiment, we will designate each patient as the target task, while using the other $n=41$ patients as source tasks. In particular, we will take $N_i=30$, and hence $N=1230$, and again since the total number of evaluations is large, will focus on BLR. The results obtained by averaging over different patients as the target task ($20$ runs per task) are shown in Figure \ref{fig:patient}. On this dataset, we observe similar behaviour of transfer methods which were able to leverage the source task information and for many patients few-shot the optimum. This suggests the presence of similar source tasks in practice and that this similarity can be exploited in the context of hyperparameter learning.
\subsection{Classification: protein dataset}
\label{app:protein}

\textbf{Jaccard kernel C-SVM} \\
Hyperparameters: $C \in [2.0^{-7}, 2.0^{10}]$ \\
Source task's random and BO iterations: $10, 10$ \\
Target task's noneBO random and BO iterations: $9, 11$

\begin{figure}[ht!]
    \centering
    \includegraphics[scale=0.55]{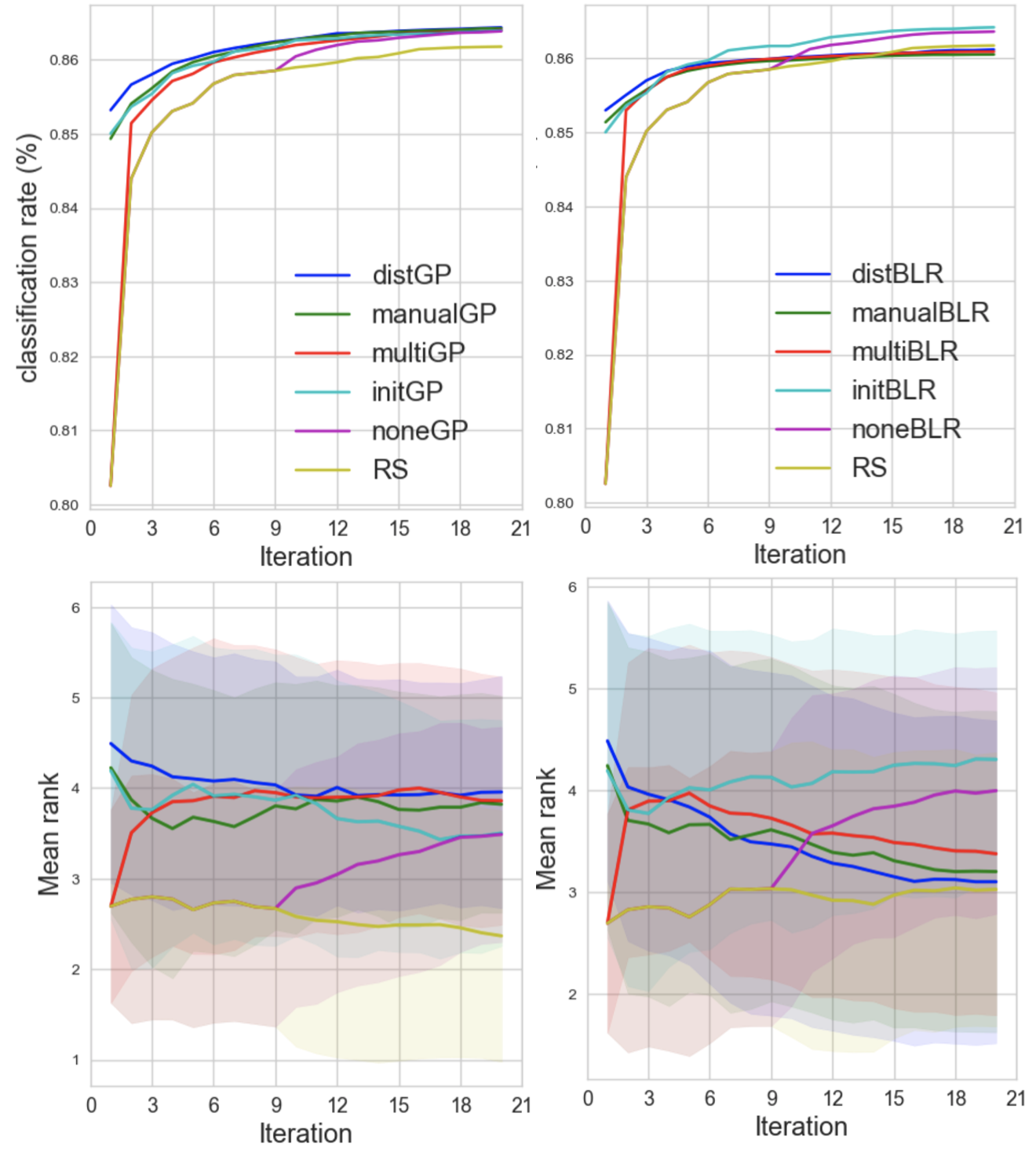}
    \caption{Protein dataset with Jaccard kernel C-SVM. Each evaluation here is averaged over $140$ runs, with each of the $7$ protein set as the target task (20 runs each). GP methods are displayed on the left, while BLR methods are displayed on the right. \textbf{Top row:} \textit{Maximum observed} classification accuracy $(\%)$. \textbf{Bottom row:} Mean rank (with respect to each run) of the different methodologies, with $\pm1$ sample standard deviation.}
    \label{fig:protein_svm}
\end{figure}

To compute the Jaccard kernel \cite{bouchard2013proof, ralaivola2005graph}, we use of the python package \textit{SciPy}\footnote{https://docs.scipy.org/doc/scipy/reference/generated/scipy.spatial.distance.cdist.html} \cite{scipy} to compute the Jaccard distance, before performing a one subtract each entry to get a similarity matrix. Results are shown in Figure \ref{fig:protein_svm}.
\pagebreak

\textbf{Random Forest} \\
Hyperparameters: \\ 
Number of trees: $n\_trees \in \{1, \dots, 200\}$ \\
Max depth of the tree: $max\_depth \in \{1, \dots, 32 \}$ \\
Min samples required to split a node (after multiplied with $s_i$):
$min\_samples\_split \in [0.01, 1.0]$ \\
Min samples required at a leaf node (after multiplied with $s_i$):
$min\_samples\_leaf \in [0.01, 0.5]$ 

Source task's random and BO iterations: $10, 10$ \\
Target task's noneBO random and BO iterations: $9, 11$

\begin{figure}[ht!]
    \centering
    \includegraphics[scale=0.53]{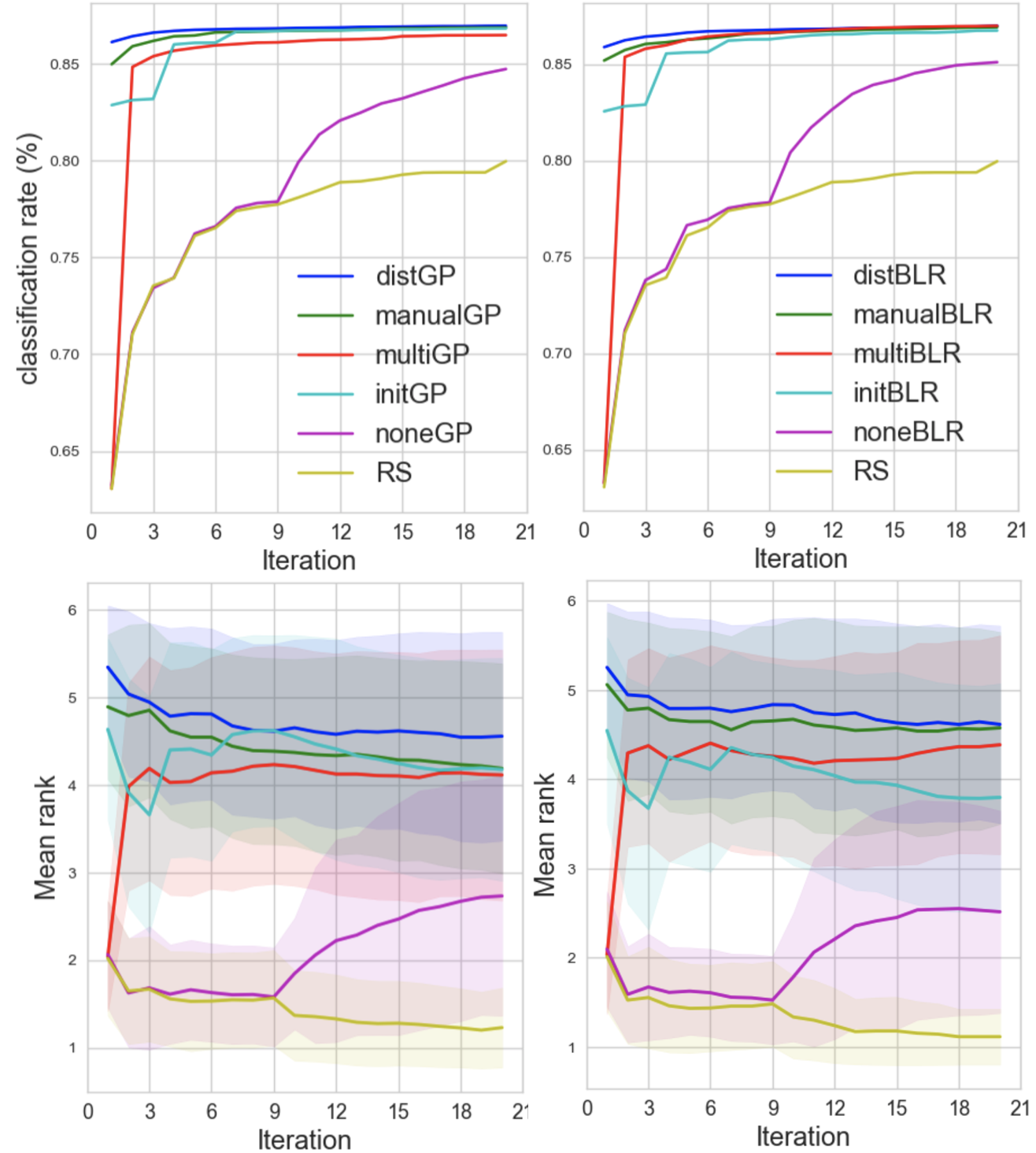}
    \caption{Protein dataset with random forest. Each evaluation here is averaged over $140$ runs, with each of the $7$ protein set as the target task (20 runs each). GP methods are displayed on the left, while BLR methods are displayed on the right. \textbf{Top row:} \textit{Maximum observed} classification accuracy $(\%)$. \textbf{Bottom row:} Mean rank (with respect to each run) of the different methodologies, with $\pm1$ sample standard deviation.}
    \label{fig:protein_forest}
\end{figure}

Since $n\_trees$ and $max\_depth$ are discrete hyperparameters, in practice we round up to the nearest integer, after a continuous version of it is proposed. For additional information on these hyperparameters, please refer to the \textit{RandomForestClassifier}\footnote{https://scikit-learn.org/stable/modules/generated/sklearn.ensemble.RandomForestClassifier.html} in the Python package \textit{scikit-learn} \cite{scikit-learn}. Results are shown in Figure \ref{fig:protein_forest}.

\end{document}